\documentclass[11pt]{article}
\usepackage{cl.forTRonly,latexsym,natbib,here}

\title{Word-to-Word Models of \\ Translational Equivalence}

\author{I. Dan Melamed
	\thanks{Dept. of Computer and Information Science,
Philadelphia, PA, 19104, U.S.A.} \\ \affil{University of Pennsylvania}
}

\date{}

\input{psfig}

\newcommand{\ignore}[1]{}

\begin{document}

\issue{0}{0}{0000}
\runningtitle{Models of Translational Equivalence}
\runningauthor{Melamed}

\maketitle

\begin{abstract}

Parallel texts (bitexts) have properties that distinguish them from
other kinds of parallel data.  First, most words translate to only one
other word.  Second, bitext correspondence is noisy.  This article
presents methods for biasing statistical translation models to reflect
these properties.  Analysis of the expected behavior of these biases
in the presence of sparse data predicts that they will result in more
accurate models.  The prediction is confirmed by evaluation with
respect to a gold standard --- translation models that are biased in
this fashion are significantly more accurate than a baseline
knowledge-poor model.  This article also shows how a statistical
translation model can take advantage of various kinds of pre-existing
knowledge that might be available about particular language pairs.
Even the simplest kinds of language-specific knowledge, such as the
distinction between content words and function words, is shown to
reliably boost translation model performance on some tasks.
Statistical models that are informed by pre-existing knowledge about
the model domain combine the best of both the rationalist and
empiricist traditions.

\end{abstract}

\section{Introduction}

The idea of a computer system for translating from one language to
another is almost as old as the idea of computer systems.  The
earliest written record of this idea is a 1949 memorandum by Warren
Weaver.  More recently, \citet{candide} have proposed methods for
constructing machine translation systems automatically.  Instead of
codifying the human translation process from introspection,
\citeauthor{candide} appealed to machine learning techniques to induce
models of the process from examples of its input and output.  The
proposal generated much excitement, because it held the promise of
automating a task that fifty years of research have proven extremely
labor-intensive and error-prone.  Yet, very few other researchers have
taken up the cause, partly because \citeauthor{ibm}'s approach was
quite a departure from the paradigm in vogue at the time.

Formally, \citeauthor{ibm} built statistical models of translational
equivalence (or {\bf translation models}\footnote{Note that the term
{\bf translation model}, which is standard in the literature, refers
to a static mathematical relationship between two data sets.  In this
usage, the term says nothing about the {\em process} of translation,
automated or otherwise.}, for short).  {\bf Translational equivalence}
is a relation that holds between two expressions with the same
meaning, where the two expressions are in different languages.  As
with all statistical models, the best translation models are those
whose parameters correspond best with the sources of variance in the
data.  Translation models whose parameters reflect existing knowledge
about particular languages and language pairs and/or universal
properties of translational equivalence benefit from the best of both
the empiricist and rationalist traditions.  This article presents
three such models, along with methods for efficiently estimating their
parameters.

More specifically, in this article, I introduce methods for modeling
three universal properties of translational equivalence in parallel
texts ({\bf bitexts}):
\begin{enumerate}
\item Most word tokens translate to only one word token.  I capture
this tendency in a one-to-one assumption.
\item Most text segments are not translated word-for-word.  I build an
explicit noise model.
\item Different linguistic objects have statistically different
behavior in translation.  I show a way to condition translation models
on different word classes to help account for the variety.
\end{enumerate}
Quantitative evaluation with respect to a gold standard has shown that
each of the three methods effects a significant improvement in
translation model accuracy.

A review of previously published translation models follows an
introduction to the different kinds of possible translation models.
The core of the article is a presentation of the model estimation
biases described above and an analysis of their expected behavior in
the face of sparse data.  The last section reports the results of a
variety of experiments designed to evaluate these innovations.

Throughout this article, I shall use ${\cal CALLIGRAPHIC}$ letters to
denote entire text corpora and other sets of sets, CAPITAL letters to
denote collections, including strings and bags, and {\em italics} for
scalar variables.  I shall also distinguish between {\bf types} and
tokens by using {\bf bold font} for the former and plain font for the
latter.

\section{Translation Model Decomposition}
\label{tmdecomp}

There are two kinds of applications of translation models: those where
word order plays a crucial role and those where it doesn't.
Empirically estimated models of translational equivalence among word
types can play a central role in both kinds of applications.

Applications where word order is not important (or at least not
essential) include
\begin{itemize}
\item cross-language information retrieval \citep[\protect{\em e.g.}\ ][]{clir}, 
\item computer-assisted language learning \citep{nerbonne},
\item certain machine-assisted translation tools \citep[\protect{\em
e.g.}\ ][]{transcheck,adomit},
\item concordancing for bilingual lexicography \citep{catiz,wordcorr},
\item corpus linguistics \citep[\protect{\em e.g.}\ ][]{corpling},
\item ``crummy'' machine translation on the internet \citep{crummy}.
\end{itemize}
For these applications, empirical models have a number of
advantages over hand-crafted models such as on-line versions of
printed bilingual dictionaries.  Two of the advantages are the
possibility of better coverage and the possibility of frequent
updates by non-expert users to keep up with rapidly evolving
vocabularies.  

A third advantage is that empirical models can provide more
accurate information about the relative importance of different
translations.  Such information is crucial for applications such as
cross-language information retrieval (CLIR).  In CLIR, the query
vector~$Q'$ is in a different language (a different vector space) from
the document vectors~$D$.  Matrix multiplication by a word-to-word
translation model~$T$ can map~$Q'$ into a vector~$Q$ in the vector
space of~$D$: \( Q = Q'T\).  In order for the mapping to be accurate,
$T$ must be able to encode many levels of relative importance among
the possible translations of each element of~$Q'$.  A typical
machine-readable bilingual dictionary says only what the possible
translations are, which is equivalent to positing a uniform
translational distribution.  The performance of cross-language
information retrieval with a uniform~$T$ is likely to be limited in
the same way as the performance of conventional information retrieval
without term frequency information, {\em i.e.}\ where the system knows
which terms occur in which documents, but not how often
\citep{buckley}.

Fully automatic high-quality machine translation is the prototypical
application where word order is crucial.  In such an application, a
word-to-word translation model can serve as an independent module in a
more complex string-to-string translation model.  The independence of
such a module is desirable for two reasons, one practical and one
philosophical.  The practical reason is illustrated in this article:
Order-independent translation models can be accurately estimated more
efficiently in isolation.  The philosophical reason is that words are
an important epistemological category in our naive mental
representations of language.  We have many intuitions (and even some
testable theories) about what words are and how they behave.  We can
bring these intuitions to bear on our translation models without being
distracted by other facets of language, such as phrase structure.  For
example, Chapter~9 of my dissertation is based on the intuition that
words can have multiple senses \citep{mythesis};
\citet{ibm}'s Model~3 and my work on non-compositional compounds
\citep{emnlp97} are based on the intuition that spaces in text do not
necessarily delimit words.

The independence of a word-to-word translation module in a
string-to-string translation model can be effected by a two-stage
decomposition.  The first stage is based on the observation that every
string {\bf S} is just an ordered bag, and that the bag {\bf B} can be
modeled independently of its order {\bf O}.  For example, the string
$\langle a b c \rangle$ consists of the bag $\{c, a, b\}$ and the
ordering relation $\{ (b , 2), (a , 1), (c , 3) \}$.  If we represent
each string {\bf S} as a pair {\bf (B,~O)}, then
\begin{eqnarray}
\Pr({\bf S}) & \equiv & \Pr({\bf B, O}) \\
	& = & \Pr({\bf B}) \cdot \Pr({\bf O | B}) .
\end{eqnarray}
Now, let ${\bf S_1}$ and ${\bf S_2}$ be two strings and let
{\bf A} be a one-to-one mapping between the elements of ${\bf S_1}$
and the elements of ${\bf S_2}$. Borrowing a term from the operations
research literature, I shall refer to such mappings as {\bf
assignments}\footnote{Assignments are different from \citet{ibm}'s
alignments in that assignments can range over pairs of arbitrary
labels, not necessarily string position indexes.  Also, unlike
alignments, assignments must be one-to-one.}.  Let ${\cal A}$ be the
the set of all possible assignments between ${\bf S_1}$ and ${\bf
S_2}$.  Using assignments, we can decompose conditional and joint
probabilities over strings:
\begin{eqnarray}
\Pr({\bf S_1 | S_2}) & = & \sum_{A \in {\cal A}} \Pr({\bf S_1, A | S_2}) \\
\Pr({\bf S_1 , S_2}) & = & \sum_{A \in {\cal A}} \Pr({\bf S_1, A, S_2})
\end{eqnarray}
where
\begin{eqnarray}
\Pr({\bf S_1, A | S_2}) & \equiv & \Pr({\bf B_1, O_1, A | S_2}) \\
	& = & \Pr({\bf B_1, A | S_2}) \cdot \Pr({\bf O_1 | B_1, A, S_2}) \\
\Pr({\bf S_1 , A, S_2}) & \equiv & \Pr({\bf B_1, O_1, A, B_2, O_2}) \\
	& = & \Pr({\bf B_1 , A, B_2}) \cdot \Pr({\bf O_1, O_2 | B_1,
	A, B_2})
\end{eqnarray}

The second stage of decomposition takes us from bags of words to the
words that they contain.  The following bag-pair generation process
illustrates how a word-to-word translation model can be embedded in a
bag-to-bag translation model for languages $L1$ and $L2$:
\begin{enumerate}
\item Generate a bag size $b$ with probability $Z(b)$ (mnemonic: $Z$
is the $siZe$ distribution).  $b$ is also the assignment size.
\item Generate $b$ language-independent concepts $C_1, \ldots, C_b$.
\item From each concept $C_i$, $1 \leq i \leq b$, generate a pair of
word strings $(\vec{ u}_i, \vec{ v}_i)$ from \mbox{$L1^* \times
L2^*$}, according to the distribution $trans(\vec{\bf u}, \vec{\bf
v})$, to lexicalize the concept in the two languages.  Some concepts
are not lexicalized in some languages, so one of $\vec{ u}_i$ and
$\vec{ v}_i$ may be empty.
\end{enumerate}
A pair of bags containing $m$ and $n$ non-empty word strings can be
generated by a process where $b$ is anywhere between 1 and $m + n$.

Without loss of generality, we can assume that each different pair of
word string types $(\vec{\bf u}, \vec{\bf v})$ is deterministically
generated from a different concept.  Thus, a bag-to-bag translation
model can be fully specified by the distributions $Z$ and $trans$.
For notational convenience, the elements of the two bags can be
labeled so that ${\bf B_1} \equiv \{ {\bf \vec{u_1}, \ldots,
\vec{u_b}} \}$ and ${\bf B_2} \equiv \{ {\bf \vec{v_1}, \ldots,
\vec{v_b}} \}$, where some of the ${\bf \vec{u}}$'s and ${\bf
\vec{v}}$'s may be empty.  The elements of an assignment, then, are
pairs of bag element labels: ${\bf A} \equiv \{ (i_1, j_1), \ldots,
(i_b, j_b) \}$, where each $i$ ranges over $\{ {\bf \vec{u_1}, \ldots,
\vec{u_b}} \}$, each $j$ ranges over $\{ {\bf \vec{v_1}, \ldots,
\vec{v_b}} \}$, each {\bf $i$} is distinct and each {\bf $j$} is
distinct.  The label pairs in a given assignment can be generated in
any order, so there are $b!$ ways to generate an assignment of size
$b$.\footnote{The number of permutations is smaller when either bag
contains two or more identical elements, but this detail will not
affect the estimation algorithms presented here.}  It follows that the
probability of generating a pair of bags $({\bf B_1, B_2)}$ with a
particular assignment {\bf A} of size $b$ is
\begin{equation}
\label{bagpairwitha}
\Pr ({\bf B_1, A, B_2} | Z, trans) = Z(b) \cdot b! {\displaystyle
\prod_{(i, j) \in {\bf A}}} trans({\bf \vec{u}_i, \vec{v}_j})
\end{equation}
The joint probability distribution $trans(\vec{\bf u}, \vec{\bf v})$
is a {\bf word-to-word translation model}.

\section{The One-to-One Assumption}
\label{121}

The most general word-to-word translation model $trans(\vec{\bf u},
\vec{\bf v})$, where $\vec{\bf u}$ and $\vec{\bf v}$ range over the
strings of $L1$ and $L2$, has an infinite number of parameters.  This
model can be constrained in various ways to make it more practical.
The models presented in this article are based on the {\bf
one-to-one~assumption}: Each word is translated to at most one other
word.  In these models, $\vec{\bf u}$ and $\vec{\bf v}$ may consist of
at most one word each.  As before, one of the two strings (but not
both) may be empty.  I shall describe empty strings as consisting of a
special {\bf {\sc null}} word, so that each word string will contain
exactly one word and can be treated as a scalar.  Henceforth, I shall
write ${\bf u}$ and ${\bf v}$ instead of $\vec{\bf u}$ and $\vec{\bf
v}$.  Under the one-to-one assumption, a pair of bags containing $m$
and $n$ non-empty words can be generated by a process where the bag
size $b$ is anywhere between $\max(m,n)$ and $m + n$.  

The one-to-one assumption is not as restrictive as it may appear: The
explanatory power of a model based on this assumption may be raised to
an arbitrary level by redefining what words are.  For example, I have
shown elsewhere how to efficiently estimate word-to-word translation
models where a word can be a non-compositional compound consisting of
several space-delimited tokens \citep{emnlp97}.  For the purposes of
this article, however, {\bf words} are the tokens generated by my
tokenizers and stemmers for the languages in question.  Therefore, the
models in this article are only a first approximation to the vast
complexities of translational equivalence.  They are intended mainly
as stepping stones towards better models.

\section{Previous Work}
\label{prevwork}

Most methods for estimating translation models from bitexts start with
the following intuition: Words that are translations of each other are
more likely to appear in corresponding bitext regions than other pairs
of words.  Following this intuition, most authors begin by counting
the number of times that word types in one half of the bitext co-occur
with word types in the other half.  Different co-occurrence counting
methods stem from different models of co-occurrence.  

A {\bf model of co-occurrence} is a boolean predicate, which indicates
whether a given pair of word {\em tokens} co-occur in corresponding
regions of the bitext space.  Different models of co-occurrence are
possible, depending on the kind of bitext map that is available, the
language-specific information that is available, and the assumptions
made about the nature of translational equivalence.  All the
translation models reviewed and introduced in this article can be
based on any of the co-occurrence models described by \citet{coocmod}.
For expository purposes, however, I shall assume a boundary-based
model of co-occurrence throughout this article.  A boundary-based
model of co-occurrence assumes that both halves of the bitext have
been segmented into $s$ segments, so that segment $U_i$ in one half of
the bitext and segment $V_i$ in the other half are mutual
translations, $1 \leq i \leq s$.  Under this model of co-occurrence,
the co-occurrence count $cooc({\bf u,v})$ for word types {\bf u} and
{\bf v} is the number of times that ${\bf u} \in U_i$ and ${\bf v} \in
V_i$ in some aligned segment pair $i$.

\subsection{Non-Probabilistic Translation Lexicons}
\label{firstord}

Many researchers have proposed greedy algorithms for estimating
non-probabilistic word-to-word translation models, also known as
translation lexicons \citep[\protect{\em e.g.}\
][]{catiz,wordcorr,funb,kumano,mel95,wuxia}.  Most of these
algorithms can be summarized as follows:
\begin{enumerate}
\item Choose a similarity function~$S$ between word types in $L1$ and
word types in $L2$.
\item Compute association scores $S({\bf u,v})$ for a set of word type
pairs $({\bf u,v}) \in (L1 \times L2)$ that occur in training data.
\item Sort the word pairs in descending order of their association
scores.
\item Discard all word pairs for which $S({\bf u,v})$ is less than a chosen
threshold $t$.  The remaining word pairs become the entries in the
translation lexicon.
\end{enumerate}
The various proposals differ mainly in their choice of similarity
function.  Almost all the similarity functions in the literature are
based on a model of co-occurrence with some linguistically-motivated
filtering \citep[see][for a notable exception]{funb}.

\begin{figure}[htb]
\centerline{\psfig{figure=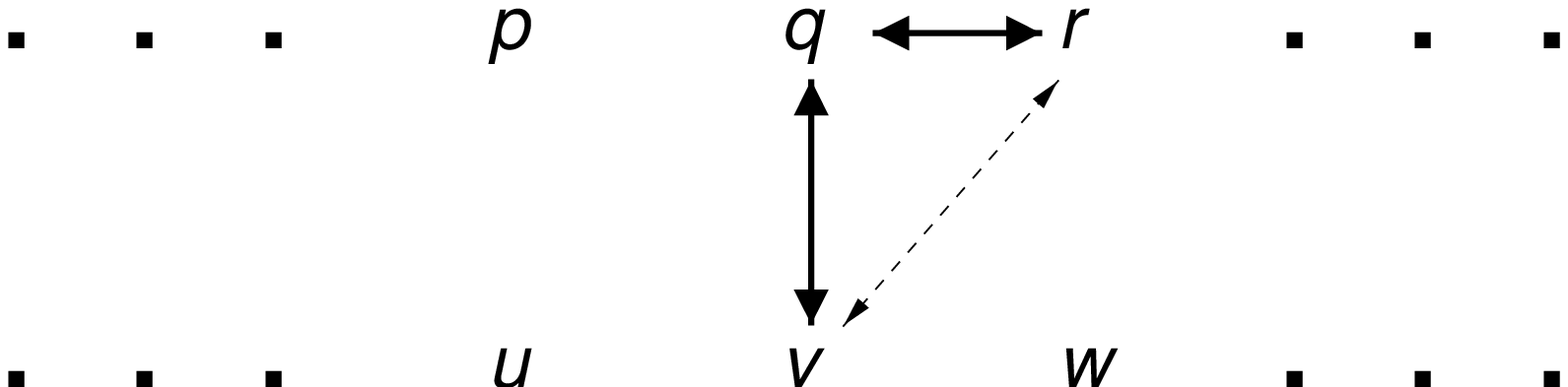,width=4.5in}}
\caption[{\em Indirect association}]{{\em $q$ and $v$ often co-occur,
as do $q$ and~$r$.  The direct association between $q$ and $v$, and
the direct association between $q$ and $r$ give rise to an indirect
association between $v$ and~$r$.}\label{dep}}
\end{figure}

Given a reasonable similarity function, the greedy algorithm works
remarkably well, considering how simple it is.  However, the
association scores in Step~2 are typically computed independently of
each other. The problem with this independence assumption is
illustrated in Figure~\ref{dep}.
The two strings represent corresponding regions of a bitext.  If ${\bf
q}$ and ${\bf v}$ co-occur much more often than expected by chance,
then any reasonable similarity metric will deem them likely to be
mutual translations.  If ${\bf q}$ and ${\bf v}$ are indeed mutual
translations, then their tendency to co-occur is called a {\bf direct
association}.  Now, suppose that ${\bf q}$ and ${\bf r}$ often
co-occur within their language.  Then ${\bf v}$ and ${\bf r}$ will
also co-occur more often than expected by chance.  The arrow between
$v$ and $r$ in Figure~\ref{dep} represents an {\bf indirect
association}, since the association between $v$ and $r$ arises only by
virtue of the association between each of them and $q$.  Models of
translational equivalence that are ignorant of indirect associations
have ``a tendency \ldots to be confused by collocates''
\citep{wordalign}.

Paradoxically, the irregularities (noise) in text and in translation
mitigate the problem.  If noise in the data reduces the strength of a
direct association, then the same noise will reduce the strengths of
any indirect associations that are based on this direct association.
On the other hand, noise can reduce the strength of an indirect
association without affecting any direct associations.  Therefore,
direct associations are usually stronger than indirect associations.
If all the entries in a translation lexicon are sorted by their
association scores, the direct associations will be very dense near
the top of the list, and sparser towards the bottom.

\citet{wordcorr} have shown that entries at the very top of the list
can be over 98\% correct.  Their algorithm gleaned lexicon entries for
about 61\% of the word tokens in a sample of 800 English sentences.
To obtain 98\% precision, their algorithm selected only entries for
which it had high confidence that the association score was high.
These would be the word pairs that co-occur most frequently.  A random
sample of 800 sentences from the same corpus showed that 61\% of the
word tokens, where the tokens are of the most frequent types,
represent 4.5\% of all the word types.  A similar strategy was
employed by \citet{wuxia} and by \citet{funb}.\footnote{These two
results should not be judged on the same scale, because it is arguably
more difficult to construct translation lexicons between English and
Chinese than between English and French.}  Fung skimmed off the top
23.8\% of the noun-noun entries in her lexicon to achieve a precision
of 71.6\%.  Wu~\&~Xia have reported automatic acquisition of 6517
lexicon entries from a 3.3-million-word corpus, with a precision of
86\%.  The first 3.3 million word tokens in an English corpus from a
similar genre contained 33490 different word types, suggesting a
recall of roughly 19\%.  Note, however, that Wu~\&~Xia chose to weight
their precision estimates by the probabilities attached to each entry:
\begin{quote}
For example, if the translation set for English word {\em detect} has
the two correct Chinese candidates with 0.533 probability and with
0.277 probability, and the incorrect translation with 0.190
probability, then we count this as 0.810 correct translations and
0.190 incorrect translations. \citep[p.\ 211]{wuxia}
\end{quote}
This is a reasonable evaluation method, but it is not comparable to
methods that simply count each lexicon entry as either right or wrong
\citep[\protect{\em e.g.}\ ][]{daille,amta}.  A weighted precision
estimate pays more attention to entries that are more frequent and
hence easier to estimate.  Therefore, weighted precision estimates are
generally higher than unweighted ones.

\subsection{Re-estimated String-to-String Translation Models}
\label{string2string}

Most translation model re-estimation algorithms published to date are
variations on the theme proposed by \citet{ibm}.  These models involve
conditional probabilities, but they can be compared to symmetric
models based on joint probabilities if the latter are normalized by
the appropriate marginal distribution.  I shall review these models
using the notation in Table~\ref{TMnotation}.
\begin{table*}[htb]
\centering
\begin{tabular}{lcl}
\hline
$({\cal U},{\cal V})$ & = & the two halves of the bitext \\
$ (U,V) $ & = & a pair of aligned text segments in $({\cal U},{\cal V})$ \\
$e({\bf u})$ & = & the frequency of ${\bf u}$ in $U$ \\
$f({\bf v})$ & = & the frequency of ${\bf v}$ in $V$ \\
$ cooc({\bf u, v}) $ & = & the number of times that ${\bf u}$ and ${\bf
v}$ co-occur \\
$ links({\bf u, v}) $ & = & the number of times that ${\bf u}$ and ${\bf
v}$ are hypothesized to \\
& & co-occur as mutual translations \\
$ trans({\bf v | u}) $ & = & the probability that a token of ${\bf u}$
will be translated as \\
& & a token of ${\bf v}$ \\
\hline
\end{tabular}
\caption{{\em Variables used to describe translation
models.}\label{TMnotation}}
\end{table*}

Methods for estimating translation parameters from co-occurrence
counts invariably involve {\bf link counts} $links({\bf u,v})$, which
represent hypotheses about the number of times that {\bf u} and {\bf
v} were generated together from the same language-independent concept,
for each {\bf u} and {\bf v} in the bitext.  A {\bf link token} is an
ordered pair of word tokens, one from each half of the bitext.  A {\bf
link type} is an ordered pair of word types.  The link counts
$links({\bf u,v})$ range over link types.  

\subsubsection{Models Using Only Co-occurrence Information}
\label{cooconly}
Brown~{\em et~al.}'s Model~1 is estimated from co-occurrence
information only, using the Expectation-Maximization (EM) algorithm
\citep{em}.

\noindent E~step:
\begin{equation}
\label{kfe}
links({\bf u, v}) = \sum_{(U,V) \in ({\cal U},{\cal V})}
	\frac{trans({\bf v | u})} {\sum_{u' \in U} trans({\bf v |
	u'})} e({\bf u}) \cdot f({\bf v})
\end{equation}
\linebreak
M~step:
\begin{equation}
\label{tfe}
trans({\bf v | u}) = \frac{links({\bf u, v})}{\sum_{\bf u'} links({\bf
u', v})}
\end{equation}
It is instructive to consider the form of Equation~\ref{kfe} when all
the translation probabilities $trans({\bf v | u})$ for a particular
${\bf u}$ are initialized to the same constant $p$, as \citet[p.\
273]{ibm} actually do:
\begin{eqnarray}
links({\bf u, v}) & = & \sum_{(U,V) \in ({\cal U},{\cal V})}
	\frac{p \cdot e({\bf u}) \cdot f({\bf v})}{p \cdot length(U)} \\ 
	& = & \sum_{(U,V) \in ({\cal U},{\cal V})} \frac{e({\bf u})
	\cdot f({\bf v})}{length(U)}
\label{kfeconst}
\end{eqnarray}
The initial link count for each ${\bf (u, v)}$ pair is set proportional to
the co-occurrence count of ${\bf u}$ and ${\bf v}$ and inversely
proportional to the length of each segment $U$ in which ${\bf u}$
occurs.  The intuition behind the numerator is central to most
bitext-based translation models: The more often two words co-occur,
the more likely they are to be mutual translations.  The intuition
behind the denominator is that the co-occurrence count of ${\bf u}$
and ${\bf v}$ should be discounted to the degree that ${\bf v}$ also
co-occurs with other words in the same segment pair.

Now consider how Equation~\ref{kfeconst} would behave under a
distance-based model of co-occurrence \citep{coocmod}, where each
token of ${\bf v}$ co-occurs with exactly $c$ words (where $c$ is
constant):
\begin{eqnarray}
links({\bf u, v}) & = & \sum_{(U,V) \in ({\cal U},{\cal V})}
				\frac{e({\bf u}) \cdot f({\bf v})}{c} \\
 	& = &  \frac{1}{c} \sum_{(U,V) \in ({\cal U},{\cal V})} e({\bf u}) \cdot f({\bf v})
\label{kfedist}
\end{eqnarray}
The discount factor $\frac{1}{c}$ disappears in the M~step. The only
difference between Equations~\ref{kfeconst} and~\ref{kfedist} is that
the former discounts co-occurrences proportionally to the segment
lengths.  When information about segment lengths is not available,
Model~1's initial parameters boil down to co-occurrence counts.

\subsubsection{Word Order Correlation Biases}
\label{ordcorrbias}
In any bitext, the positions of words with respect to the true bitext
map correlate with the positions of their translations.  The
correlation is stronger for language pairs with more similar word order.
\citet{candide} introduced the idea that this correlation can be
encoded in translation model parameters.  \citet{wordalign}\ expanded
on this idea by replacing \citeauthor{ibm}'s word alignment
parameters, which were based on absolute word positions in aligned
segments, with a much smaller set of relative offset parameters.  The
much smaller number of parameters allowed \citeauthor{wordalign}'s
model to be effectively trained on much smaller bitexts.
\citet{vogel} have shown how some additional independence assumptions can
turn this model into a Hidden Markov Model, enabling even more
efficient parameter estimation.

It cannot be overemphasized that the word order correlation bias is
just knowledge about the problem domain, which can be used to guide
the search for the optimum model parameters.  Translational
equivalence can be empirically modeled for any pair of languages, but
some models and model biases work better for some language pairs than
for others.  The word order correlation bias is most useful when it
has high predictive power, {\em i.e.}\ when the distribution of
alignments or offsets has low entropy.  The entropy of this
distribution is indeed relatively low for the language pair that both
Brown~{\em et~al.}\ and Dagan~{\em et~al.}\ were working with ---
French and English have very similar word order.  A word order
correlation bias would be of less benefit with noisier training
bitexts or for language pairs with less similar word order.  The same
is true of the phrase structure biases in \citet{ibm}'s Models~4 and~5.

\subsection{Re-estimated Bag-to-Bag Translation Models}

At about the same time that I developed the models in this article,
\citet{hiemstra} independently developed his own bag-to-bag model of
translational equivalence.  His model is also based on a one-to-one
assumption, but it differs from my models in that it does not allow
empty words to be generated.  {\em I.e.}, it assumes that the two bags
in each pair contain the same number of words.  His estimation method
is also different in that it does not impose any structure on the
hidden parameters: Whereas my estimation methods revolve around
assignments (see Equation~\ref{bagpairwitha}), \citeauthor{hiemstra}
has modeled pairs of word bags as a multinomial over the crossproduct
of two vocabularies.  Maximum likelihood parameter estimation is
computationally too expensive for \citeauthor{hiemstra}'s model, so he
proposed the Iterative Proportional Fitting Procedure (IPFP)
\citep{ipfp} as a cheaper approximation method.

The IPFP is quite sensitive to initial conditions, so
\citeauthor{hiemstra} investigated a number of initialization options.
Choosing the most advantageous, \citeauthor{hiemstra} has published
parts of the translational distributions of certain words, induced
using both his method and \citet{ibm}'s Model~1 from the same training
bitext.  Subjective comparison of these examples suggests that
\citeauthor{hiemstra}'s method is more accurate.  \citet{hiemstra2}
has also evaluated the recall and precision of his method and of
Model~1 on a small hand-constructed set of link tokens in a particular
bitext.  Model~1 fared worse, on average.

\section{Parameter Estimation}
\label{paramest}

This section describes my methods for estimating the translation
parameters of a symmetric word-to-word translation model from a
bitext.  For most applications, we are interested in estimating the
probability $trans({\bf u, v})$ of generating the pair of words $({\bf
u,v})$.  For estimation purposes, however, it is more convenient to
deal with likelihoods $like({\bf u,v})$, the likelihood that ${\bf u}$
and ${\bf v}$ can ever be mutual translations, {\em i.e.}\ that there
exists some context where tokens $u$ and $v$ are generated from the
same concept.  There are various possible definitions for $like({\bf
u,v})$ and the relationship between $like({\bf u,v})$ and $trans({\bf
u, v})$ can be more or less direct, depending on the model.  The
maximum likelihood estimate of $trans({\bf u, v})$ can always be
derived by normalizing the link counts so that $\sum_{\bf u,v}
trans({\bf u, v}) = 1$:
\begin{equation}
\label{mletrans}
trans({\bf u, v}) = \frac{links({\bf u,v})}{\sum_{\bf u',v'} links({\bf
u',v'})}
\end{equation}

Link counts, and therefore also the translation parameters $trans({\bf
u, v})$, cannot be directly observed in a training bitext, because we
don't know which words in one half of the bitext were generated
together with which words in the other half.  The observable features
of the bitext are only the co-occurrence counts $cooc({\bf u,v})$ (see
Section~\ref{prevwork}).  All my methods for estimating the
translation parameters $trans({\bf u, v})$ from the co-occurrence
counts $cooc({\bf u,v})$ share the following general outline:
\begin{enumerate}
\item Initialize the model parameters to a first approximation.
\item Estimate the link counts $links({\bf u,v})$, as a function of the
model parameters and the co-occurrence counts.
\item Estimate the model parameters $like({\bf u,v})$, as a function of
the link counts and the co-occurrence counts.
\item Repeat from Step 2, until the model converges to the desired
degree.  I have adopted the simple heuristic that the model has converged
when less than .0001 of the $trans({\bf u, v})$ distribution changes
from one iteration to the next.
\item Compute the maximum likelihood estimate (MLE) of $trans({\bf u,
v})$, by normalizing the converged link counts as in Equation~\ref{mletrans}.
\end{enumerate}
Under certain conditions, a parameter estimation process of this sort
is an instance of the Expectation-Maximization (EM) algorithm
\citep{em}.  As explained below, meeting these conditions is
computationally too expensive for my models.  Therefore, I employ some
approximations, which lack the EM algorithm's convergence guarantee.

The maximum likelihood approach to estimating the unknown
parameters is to find the set of parameters $\widehat{\Theta}$ that
maximize the probability of the training bitext $(U,V)$.
\begin{equation}
\label{thetamax}
\widehat{\Theta} =  \arg \max_{\Theta} \Pr(U, V| \Theta)
\end{equation}
where the probability of the bitext is a weighted sum over the
distribution ${\cal A}$ of possible assignments:
\begin{equation}
\label{pbitext}
\Pr ( U, V | \Theta) = {\displaystyle \sum_{ A \in {\cal A}}} \Pr ( U,
A, V | \Theta) .
\end{equation}
The MLE method is infeasible, because the number of possible
assignments grows exponentially with the size of the bitext.  Due to
the parameter interdependencies introduced by the one-to-one
assumption, we cannot decompose the assignments into parameters that
can be estimated independently of each other \citep[as in][Equation
26]{ibm}.  This is why we must make do with approximations to the EM
algorithm.

In this situation, \citet[p.\ 293]{ibm} recommend ``evaluating the
expectations using only a single, probable alignment.'' The single
most probable assignment $A_{max}$ is called the {\bf Viterbi
assignment}:
\begin{eqnarray}
\label{Amax}
A_{max} & = & \arg \max_{A \in {\cal A}} \Pr(U, A, V | \Theta) \\
	& = & \arg \max_{A \in {\cal A}} Z(b) \cdot b! {\displaystyle
	\prod_{(x, y) \in A}} trans({\bf u}_x, {\bf v}_y) \\
	& = & \arg \max_{A \in {\cal A}} \log \left[ Z(b) \cdot b!
	{\displaystyle \prod_{(x, y) \in A}} trans({\bf u}_x, {\bf v}_y) \right] \\
	& = &  \arg \max_{A \in {\cal A}} \left\{ \log [Z(b) \cdot b!] +
	{\displaystyle \sum_{(x, y) \in A}} \log trans({\bf u}_x, {\bf v}_y)\right\}
\end{eqnarray}
To simplify things further, let us assume that $Z(b) \cdot b!$ is
constant, so that
\begin{eqnarray}
A_{max} & = & \arg \max_{A \in
	{\cal A}} {\displaystyle \sum_{(x, y) \in A}} \log trans({\bf
	u}_x, {\bf v}_y)
\label{Amax-final}
\end{eqnarray}
If we represent the bitext as a bipartite graph and weight the edges
by $\log trans{\bf (u, v)}$, then the right-hand side of
Equation~\ref{Amax-final} is an instance of the {\em weighted maximum
matching problem} and $A_{max}$ is its solution.  For a bipartite
graph $G = (V_1 \cup V_2, E)$, with $v = |V_1 \cup V_2|$ and $e =
|E|$, the lowest currently known upper bound on the computational
complexity of this problem is $O(ve + v^2\log v)$ \citep[p.\
500]{netflow}.  Although this upper bound is polynomial, it is still
too expensive for typical bitexts.  The next subsection describes a
greedy approximation to the Viterbi approximation.

\subsection{Method A: The Competitive Linking Algorithm}
\label{MethodA}

\subsubsection{Step 1: Initialization}
Almost every published translation model estimation algorithm exploits
the well-known correlation between the link likelihoods $like({\bf u,
v})$ and the co-occurrence counts $cooc({\bf u,v})$.  As discussed in
Section~\ref{prevwork}, many algorithms also normalize the
correlation by the marginal frequencies of {\bf u} and {\bf v}.
However, these quantities account for only three of the cells in
the following contingency table: 

\begin{center}
\begin{tabular}{r||c|c||c}
		& {\bf u} &  ${\bf \neg u}$ & Total \\ \hline \hline
${\bf v}$	& $cooc({\bf u,v})$ & $cooc({\bf \neg u, v})$ 	& $cooc({\bf \cdot,  v})$ \\ \hline
${\bf \neg v}$ 	& $cooc({\bf u, \neg v})$ & $cooc({\bf \neg u, \neg
v})$ 	& $cooc({\bf \cdot, \neg v})$ \\ \hline \hline
Total		&  $cooc({\bf u, \cdot})$ &  $cooc({\bf \neg u,
\cdot})$ & $cooc({\bf \cdot , \cdot })$   \\ \hline
\end{tabular}
\end{center}
\noindent The statistical interdependence between two word types can be
estimated more robustly by considering the whole table.  For example,
\citet{wordcorr} suggest that ``$\phi^2$, a $\chi^2$-like statistic,
seems to be a particularly good choice because it makes good use of
the off-diagonal cells'' of the contingency table.  In informal
experiments reported elsewhere \citep{mel95}, I found that the $G^2$
statistic suggested by \citet{dunn} slightly 
outperforms $\phi^2$. Let the cells of the contingency table be named
as follows: 

\begin{center}
\begin{tabular}{r||c|c|}
		& {\bf u} &  ${\bf \neg u}$ \\ \hline \hline
${\bf v}$	& $a$ & $b$ \\ \hline
${\bf \neg v}$ 	& $c$ & $d$ \\ \hline
\end{tabular}
\end{center}
\noindent Now,
\begin{equation}
\label{g2}
G^2({\bf u,v}) =  - 2 \log \frac{B(a | a + b, p_1) B(c | c + d, p_2)}
				{B(a | a + b, p) B(c | c + d, p)}
\end{equation}
where $B(k | n, p) = \left( \mbox{\small $ \begin{array}{c} n \\ k
\end{array}$} \right) p^k (1-p)^{n-k}$ are binomial probabilities.  The
statistic uses maximum likelihood estimates for the probability
parameters: \mbox{\( p_1 = \frac{a}{a+b} \)}, \mbox{\( p_2 =
\frac{c}{c+d} \)}, \mbox{\( p = \frac{a + b}{a + b + c + d} \)}.
$G^2$ is easy to compute because the binomial coefficients cancel out.
All my methods initialize the model parameters $like{\bf (u, v)}$ to
$G^2{\bf (u, v)}$, except that the likelihood of any word being linked
to {\bf {\sc null}} is initialized to an infinitesimal value.  I have
also found it useful to smooth the co-occurrence counts using the
Simple Good-Turing smoothing method
\citep{goodturing} before computing~$G^2$.

\subsubsection{Step 2: Estimation of Link Counts}

To further reduce the complexity of estimating the model parameters, I
employ the {\bf competitive linking algorithm}, which is a greedy
approximation to the Viterbi approximation:
\begin{enumerate}
\item Sort all the translation likelihood estimates $like({\bf u, v)}$ from
highest to lowest.
\item For each likelihood estimate $like({\bf u, v})$, in order:
\begin{enumerate}
\item If {\bf u} (resp., {\bf v}) is {\bf {\sc null}}, consider all
tokens of {\bf v} (resp., {\bf u}) in the bitext linked to {\bf {\sc
null}}.  Otherwise, link all co-occurring token pairs $(u, v)$ in the
bitext.

\item The one-to-one assumption implies that linked words cannot be
linked again.  Therefore, remove all linked word tokens from
their respective halves of the bitext.
\end{enumerate}
\end{enumerate}
The competitive linking algorithm can be viewed as a heuristic search
for the most likely assignment in the space of all possible
assignments.  The search heuristic is that the most likely assignments
contain links that are individually the most likely.  The search
proceeds by a process of elimination.  In the first search iteration,
all the assignments that do not contain the most likely link are
discarded.  In the second iteration, all the assignments that do not
contain the second most likely link are discarded, and so on until
only one assignment remains\footnote{Given a method of assigning
probabilities to assignments, the competitive linking algorithm can be
generalized to stop searching before the number of possible
assignments is reduced to one, at which point the link counts can be
computed as weighted averages over the remaining assignments using
Equation~\ref{pbitext}.}. The algorithm greedily selects the most
likely links first, and then selects less likely links only if they
don't conflict with previous selections.  The probability of a link
being rejected increases with the number of links that are selected
before it, and thus decreases with the link's likelihood.  In this
problem domain, the competitive linking algorithm usually finds one of
the most likely assignments, as I will show in Section~\ref{TM-eval}.
Under an appropriate hashing scheme, the expected running time of the
competitive linking algorithm is linear in the size of the input
bitext.

\subsubsection{Step 3: Re-estimation of the Model Parameters}
Method~A re-estimates the model parameters simply by normalizing the
link counts to sum to 1 as in Equation~\ref{mletrans}.  The competitive
linking algorithm only cares about the {\em relative} magnitudes of
the various $like({\bf u,v})$.  However, Equation~\ref{Amax} is a sum
rather than a product, so I scale the parameters logarithmically, to
be consistent with its probabilistic interpretation:
\begin{equation}
\label{loglike}
like({\bf u,v}) = \log trans({\bf u,v})
\end{equation}

\subsection{Method B: Improved Estimation Using an Explicit Error Model}
\label{MethodB}

\citet{yar93} has shown that ``for several definitions of sense and
collocation, an ambiguous word has only one sense in a given
collocation with a probability of 90-99\%.''  In other words, a single
contextual clue can be a highly reliable indicator of a word's sense.
One of the definitions of ``sense'' studied by Yarowsky was a word
token's translation in the other half of a bitext.  For example, the
English word {\em sentence} may be considered to have two senses,
corresponding to its French translations {\em peine} (judicial
sentence) and {\em phrase} (grammatical sentence).  If a token of {\em
sentence} occurs in the vicinity of a word like {\em jury} or {\em
prison}, then it is far more likely to be translated as {\em peine}
than as {\em phrase}.  ``In the vicinity of'' is one kind of
collocation.  Co-occurrence in bitext space is another kind of
collocation.  If each word's translation is treated as a sense tag
\citep{resyar}, then ``translational'' collocations have the unique
property that the collocate and the word sense are one and the same!

\begin{figure}[htb]
\centerline{\psfig{figure=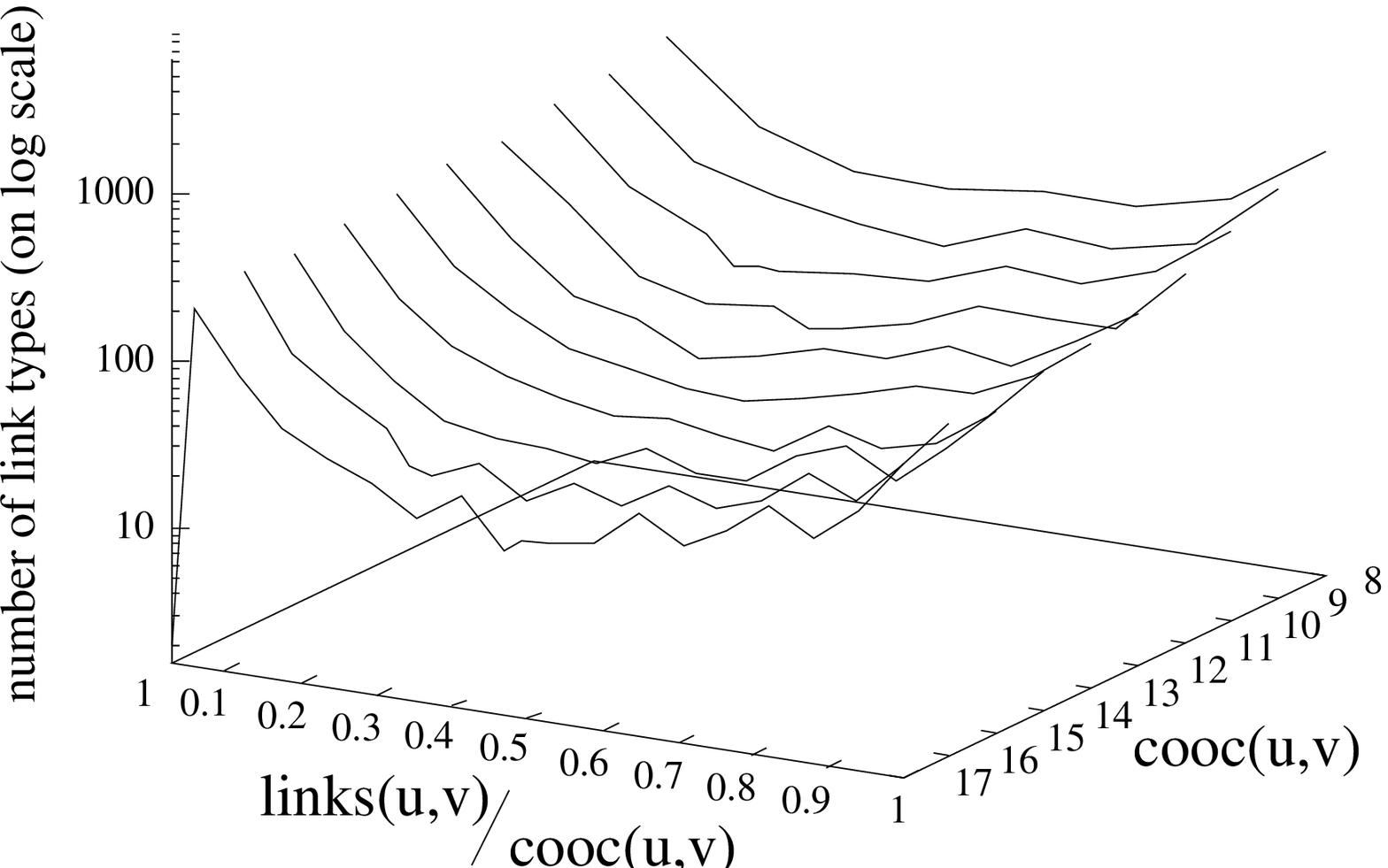,width=5in}}
\caption[{\em A fragment of the joint frequency $(\frac{links{\bf
(u,v)}}{cooc{\bf (u,v)}}, cooc{\bf (u,v)})$}]{{\em A fragment of the
joint frequency $(\frac{links{\bf (u,v)}}{cooc{\bf (u,v)}}, cooc{\bf
(u,v)})$.  Note that the frequencies are plotted on a log scale ---
the bimodality is quite sharp.}\label{bimodal}}
\end{figure}

Method~B exploits this property under the hypothesis that ``one sense
per collocation'' holds for translational collocations.  This
hypothesis implies that if ${\bf u}$ and ${\bf v}$ are {\em possible}
mutual translations, and a token $u$ co-occurs with a token $v$ in the
bitext, then with very high probability the pair $(u,v)$ was generated
from the same concept and should be linked.  To test this hypothesis,
I ran one iteration of Method~A on 300000 aligned sentence pairs from
the Canadian Hansards bitext.  I then plotted the ratio
$\frac{links{\bf (u,v)}}{cooc{\bf (u,v)}}$ for several values of
$cooc{\bf (u,v)}$ in Figure~\ref{bimodal}.
The bimodality of the surface shows that the ratio $\frac{links{\bf
(u,v)}}{cooc{\bf (u,v)}}$ tends to be either very high or very low.
Note that the frequencies are plotted on a log scale --- the
bimodality is quite sharp.

Information about how often words co-occur without being linked can be
used to bias the estimation of translation model parameters.  The
smaller the ratio $\frac{links{\bf (u,v)}}{cooc{\bf (u,v)}}$, the more
likely it is that {\bf u} and {\bf v} are {\em not} mutual
translations, and that links posited between tokens of {\bf u} and
{\bf v} are noise.  The bias can be implemented via auxiliary
parameters that model the curve illustrated in Figure~\ref{bimodal}.
The competitive linking algorithm creates all the links of a given
type independently of each other\footnote{Except for the case when
multiple tokens of the same word type occur near each other, which I
hereby sweep under the carpet.}.  So, the distribution of the number
$links{\bf (u,v)}$ of links connecting word types ${\bf u}$~and~${\bf
v}$ can be modeled by a binomial distribution with parameters
$cooc{\bf (u,v)}$ and $p{\bf (u,v)}$.  $p{\bf (u,v)}$ is the
probability that {\bf u} and {\bf v} will be linked when they
co-occur.  There is never enough data to robustly estimate each $p$
parameter separately.  Instead, I shall model the $p$'s via only two
distinct parameters.  If ${\bf u}$ and ${\bf v}$ are mutual
translations, then $p{\bf (u,v)}$ will average to a relatively high
probability, which I will call~$\lambda^+$.  If ${\bf u}$ and ${\bf
v}$ are not mutual translations, then $p{\bf (u,v)}$ will average to a
relatively low probability, which I will call~$\lambda^-$.
$\lambda^+$ and $\lambda^-$ correspond to the two peaks of the
distribution of $\frac{links{\bf (u,v)}}{cooc{\bf (u,v)}}$, a fragment
of which is illustrated in Figure~\ref{bimodal}.  The two parameters
can also be interpreted as the rates of true and false positives.  If
the translation in the bitext is consistent and the translation model
is accurate, then $\lambda^+$ will be close to~1 and $\lambda^-$ will
be close to~0.

To find the most likely values of the auxiliary parameters $\lambda^+$
and $\lambda^-$, I adopt the standard method of maximum likelihood
estimation, and find the values that maximize the probability of the
link frequency distributions, under the usual independence
assumptions, where
\begin{equation}
\label{prdata}
\Pr ( links | model ) = \prod_{{\bf u,v}} \Pr(links({\bf u,v}) |
cooc({\bf u,v}), \lambda^+, \lambda^-) .
\end{equation}
The factors on the right-hand side of Equation~\ref{prdata} can be
written explicitly with the help of a mixture coefficient.  Let $\tau$
be the probability that an arbitrary co-occurring pair of word types
are mutual translations.  Let $B(k|n,p)$ denote the probability that
$k$ links are observed out of $n$ co-occurrences, where $k$ has a
binomial distribution with parameters $n$ and $p$.  Then the
probability that two arbitrary word types ${\bf u}$ and ${\bf v}$ are
linked $links({\bf u,v})$ times out of $cooc({\bf u,v})$ co-occurrences is a
mixture of two binomials:
\begin{eqnarray}
\label{twobin}
\Pr(links({\bf u,v}) | cooc({\bf u,v}) , \lambda^+, \lambda^-) \hspace{.1in}
& = & \hspace{.3in} \tau B(links({\bf u,v}) | cooc({\bf u,v}), \lambda^+) \\
& + & (1 - \tau) B(links({\bf u,v}) | cooc({\bf u,v}), \lambda^-)
\hspace{.3in} . \nonumber
\end{eqnarray}

One more variable allows us to express $\tau$ in terms of
$\lambda^+$ and $\lambda^-$: Let $\lambda$ be the probability that an
arbitrary co-occuring pair of word tokens will be linked, regardless
of whether they are mutual translations.  Since $\tau$ is constant
over all word types, it also represents the probability that an
arbitrary co-occurring pair of word {\em tokens} are mutual
translations.  Therefore,
\begin{equation}
\label{L1}
\lambda = \tau \lambda^+ + (1 - \tau) \lambda^-.
\end{equation}
$\lambda$ can also be estimated empirically.  Let $K$ be the total
number of links in the bitext and let $N$ be the total number of
co-occuring word token pairs: 
\begin{equation}
K = \sum_{\bf u,v} links({\bf u,v}) ,
\end{equation}
\begin{equation}
\label{nsum}
N = \sum_{\bf u,v} cooc({\bf u,v}) . 
\end{equation}
By definition,
\begin{equation}
\label{L2}
\lambda = K / N.
\end{equation}
Equating the right-hand sides of Equations~\ref{L1} and~\ref{L2}
and rearranging the terms, we get:
\begin{equation}
\label{tau}
\tau = \frac{K / N - \lambda^-}{\lambda^+ - \lambda^-}.
\end{equation}
Since $\tau$ is now a function of $\lambda^+$ and $\lambda^-$, only
the latter two variables represent degrees of freedom in the model.

\begin{figure}[htb]
\centerline{\psfig{figure=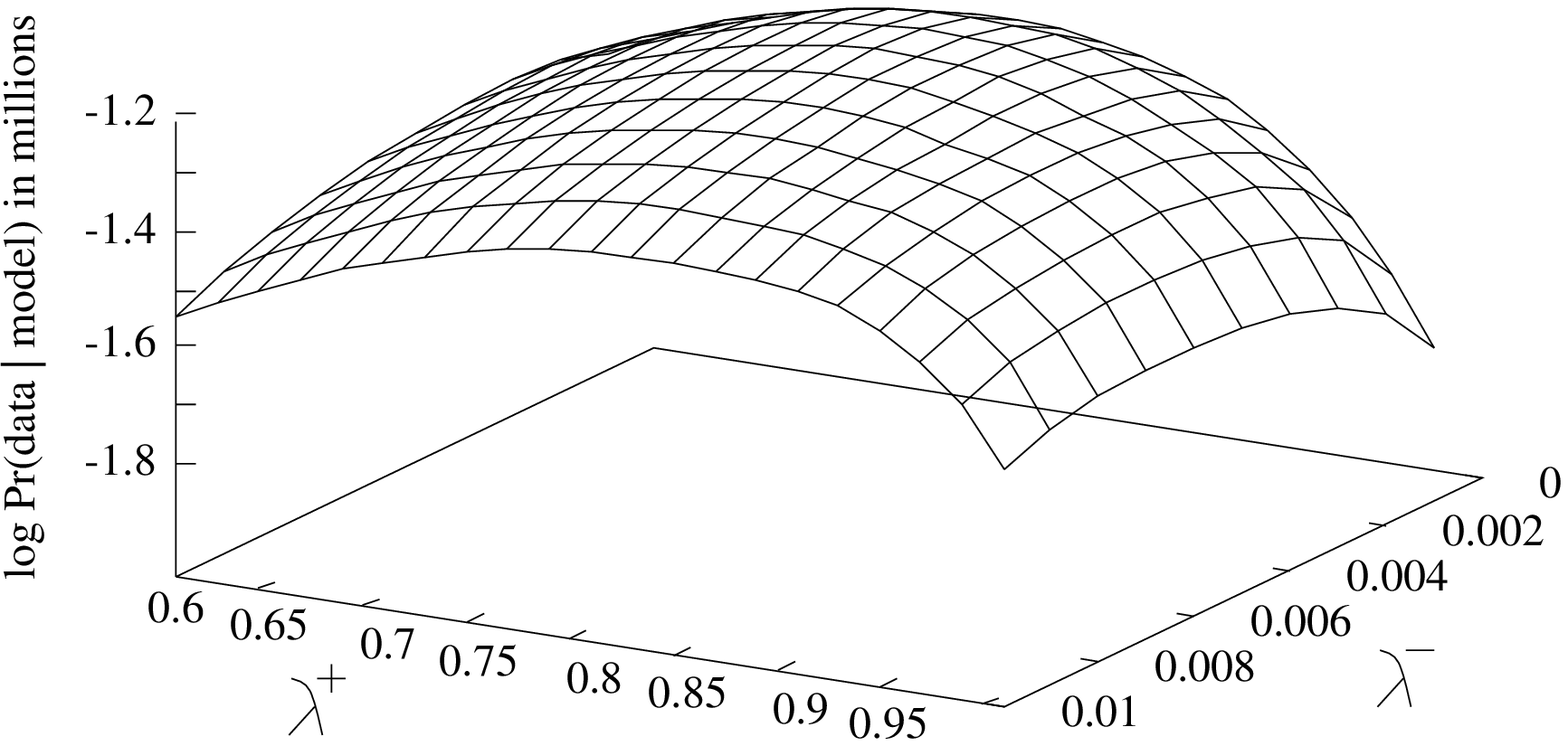,width=5in}}
\caption[{\em $\Pr(links|model)$ has only one global maximum in the
region of interest.}]{{\em $\Pr(links|model)$, as given in
Equation~\ref{prdata}, has only one global maximum in the region of
interest, where $1 > \lambda^+ > \lambda > \lambda^- >
0$.}\label{surface}}
\end{figure}
The probability function expressed by Equations~\ref{prdata}
and~\ref{twobin} may have many local maxima.  In practice, these local
maxima are like pebbles on a mountain, invisible at low resolution.  I
computed Equation~\ref{prdata} over various combinations of
$\lambda^+$ and $\lambda^-$ after one iteration over 300000 aligned
sentence pairs from the Canadian Hansard bitext.  Figure~\ref{surface}
illustrates that the region of interest in the parameter space, where
$1 > \lambda^+ > \lambda > \lambda^- > 0$, has only one dominant
global maximum.  This global maximum can be found by standard
hill-climbing methods, as long as the step size is large enough to
avoid getting stuck on the pebbles.

Given estimates for $\lambda^+$ and $\lambda^-$, we can compute
$B(links({\bf u,v}) | cooc({\bf u,v}), \lambda^+)$ and 
$B(links({\bf u,v}) | cooc({\bf u,v}), \lambda^-)$ for each occurring
combination of $links$ and $cooc$ values.  These are the probabilities that
$links({\bf u,v})$ links were generated out of $cooc({\bf u,v})$
possible links by a process that generates correct links and by a
process that generates incorrect links, respectively.  The ratio of
these probabilities is the likelihood ratio in favor of the types
${\bf u}$ and ${\bf v}$ being possible mutual translations, for all
${\bf u}$ and ${\bf v}$:
\begin{equation}
\label{ll}
like({\bf u,v}) = \log \frac{B(links({\bf u,v}) | cooc({\bf u,v}),
\lambda^+)}{B(links({\bf u,v}) | cooc({\bf u,v}), \lambda^-)}.
\end{equation}

In the preceding equations, either ${\bf u}$ or ${\bf v}$ can be {\bf
{\sc null}}.  However, the number of times that a word co-occurs with
{\bf {\sc null}} is not an observable feature of bitexts.  To make
sense of co-occurrences with {\bf {\sc null}}, we can view
co-occurrences as {\em potential} links and $cooc({\bf u,v})$ as the
maximum number of times that tokens of ${\bf u}$ and ${\bf v}$ might
be linked.  From this point of view, $cooc(${\bf u}, {\bf {\sc null}})
should be set to the marginal frequency of ${\bf u}$, since each token
of ${\bf u}$ represents one potential link to {\bf {\sc null}}.  These
co-occurrence counts should be summed together with all the others in
Equation~\ref{nsum}.

Method~B differs from Method~A only in its use of the auxiliary
parameters in Equation~\ref{ll} to re-estimate the model parameters.
These parameters and the error model that they represent can be
employed the same way in translation models that are not based on the
one-to-one assumption.  An interesting property of Equation~\ref{ll}
is that it is possible, for a given word type {\bf u}, that $like({\bf
u,v}) < 0$ for all {\bf v} including {\bf {\sc null}}.  These are the
words about which the model is uncertain, and they represent fertile
ground for future work.

\subsection{Method~C: Improved Estimation Using Pre-Existing Word Classes}
\label{MethodC}

In Method~B, the estimation of the auxiliary parameters $\lambda^+$
and $\lambda^-$ depends only on the co-occurrence counts and on the
distributions of link frequencies generated by the competitive linking
algorithm.  All word pairs that co-occur the same number of times and
are linked the same number of times are assigned the same $like$
value.  More accurate models can be induced by taking into account
various features of the linked tokens.  For example, frequent words
are translated less consistently than rare words \citep{sement}.  To
account for these differences, we can estimate separate values of
$\lambda^+$ and $\lambda^-$ for different ranges of $cooc{\bf (u,v)}$.
Similarly, the auxiliary parameters can be conditioned on the linked
parts of speech.  A kind of word order correlation bias can be
effected by conditioning the auxiliary parameters on the relative
positions of linked word tokens in their respective texts.  Just as
easily, we can model link types that coincide with entries in an
on-line bilingual dictionary separately from those that do not
\citep[\protect{\em cf.\ }][]{dictdata}.  When the auxiliary parameters
are conditioned on different link classes, Step~3 of Method~B is
repeated for each link class.

\section{Effects of Sparse Data}
\label{sparse}

The one-to-one assumption is a potent weapon against the ever-present
sparse data problem.  The assumption enables accurate estimation of
translational distributions even for words that occur only once, as
long as the surrounding words are more frequent.  In most translation
models, link likelihood is correlated with co-occurrence frequency.
So, links between tokens $u$ and $v$ for which $like({\bf u}, {\bf
v})$ is highest are the ones for which there is the most evidence, and
thus also the ones that are easiest to predict correctly.
\mbox{Winner-take-all} link assignment methods, such as the
competitive linking algorithm, can prevent links based on indirect
associations (see Section~\ref{firstord}), thereby leveraging their
accuracy on the more confident links to raise the accuracy of the less
confident links.  For example, suppose that $u_1$ and $u_2$ co-occur
with $v_1$ and $v_2$ in the training data, and the model estimates \(
like({\bf u_1}, {\bf v_1}) = .05, like({\bf u_1}, {\bf v_2}) = .02$,
and $like({\bf u_2}, {\bf v_2}) = .01 \).  According to the one-to-one
assumption, $(u_1, v_2)$ is an indirect association and the correct
translation of $v_2$ is $u_2$.  To the extent that the one-to-one
assumption is valid, it reduces the probability of spurious links for
the rarer words.  The more the incorrect candidate translations can be
eliminated for a given rare word, the more likely the correct
translation is to be found.  So, the probability of a correct match
for a rare word is proportional to the fraction of words around it
that can be linked with higher confidence.  This fraction is largely
determined by two bitext properties: the distribution of word
frequencies, and the distribution of co-occurrence counts.  I shall
explore each of these properties in turn.

The distribution of word frequencies is a function of corpus size.
The words in any text corpus are drawn from a large but finite
vocabulary.  As the corpus gets larger, fewer new words appear, and
the average frequency of words already in the corpus rises. I took
random samples of varying sizes from large text corpora in French and
in English.  The corpora comprised news text ({\em Le~Monde} and {\em
Wall~Street Journal}), parliamentary debate transcripts (Hansards) and
Sun MicroSystems software documentation (AnswerBooks).
\begin{figure}[htb]
\centerline{\psfig{figure=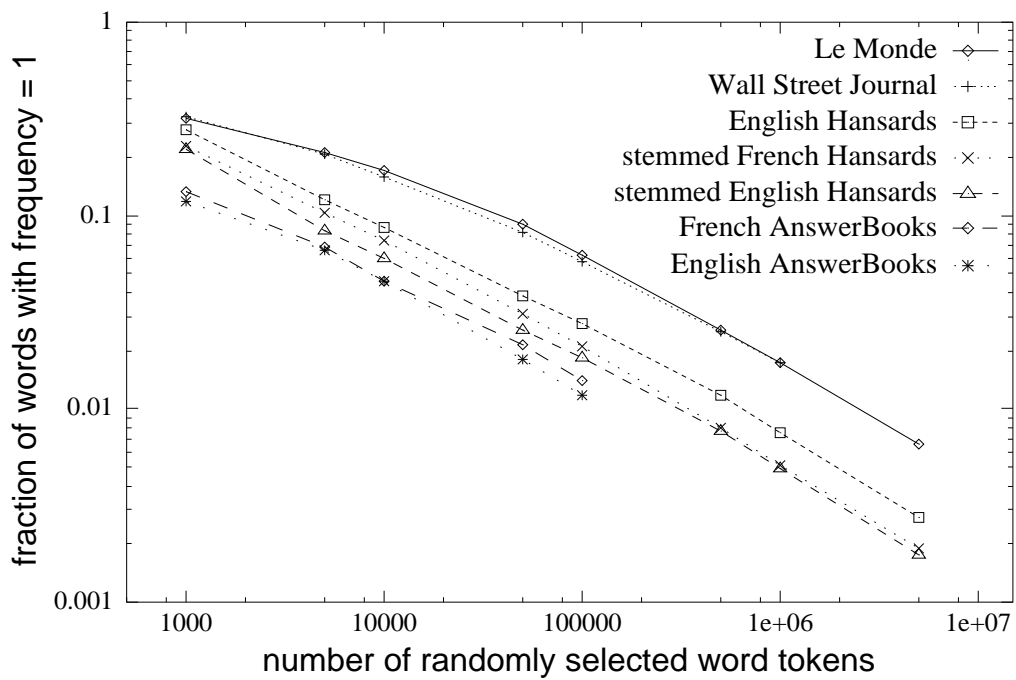,width=5.25in}}
\caption{{\em The log-log relationship between corpus size and the
proportion of singletons.}\label{singmass}}
\end{figure}
Figure~\ref{singmass} shows the log-log relationship between sample
size and the fraction of words (by token) that appear in the sample
only once.  For example, suppose we draw a random sample of one
million words from {\em Le~Monde}, and then select a random word type
{\bf w} from this random sample.  According to Figure~\ref{singmass},
the chances are roughly 0.017 that {\bf w} appears only once in that
one million words.  If the sample were only one thousand words,
however, our chances of drawing a singleton rise to 0.317.  The nearly
linear shape of the log-log curve seems largely invariant across
languages and text genres, as predicted by \citet{zipf}.  Some curves
in the graph are higher than others, because the language genres from
which the corpora were drawn have richer vocabularies.  For example,
the fraction of singleton words is consistently smaller in the stemmed
English Hansards than in the same text when it is not stemmed, which
is the whole motivation for stemming.  Figure~\ref{frqmass}, based on
{\em Le~Monde} text, shows that the log-log relationship holds for
higher frequencies too.  In a larger corpus, a larger fraction of the
word types appear more frequently.
\begin{figure}[htb]
\centerline{\psfig{figure=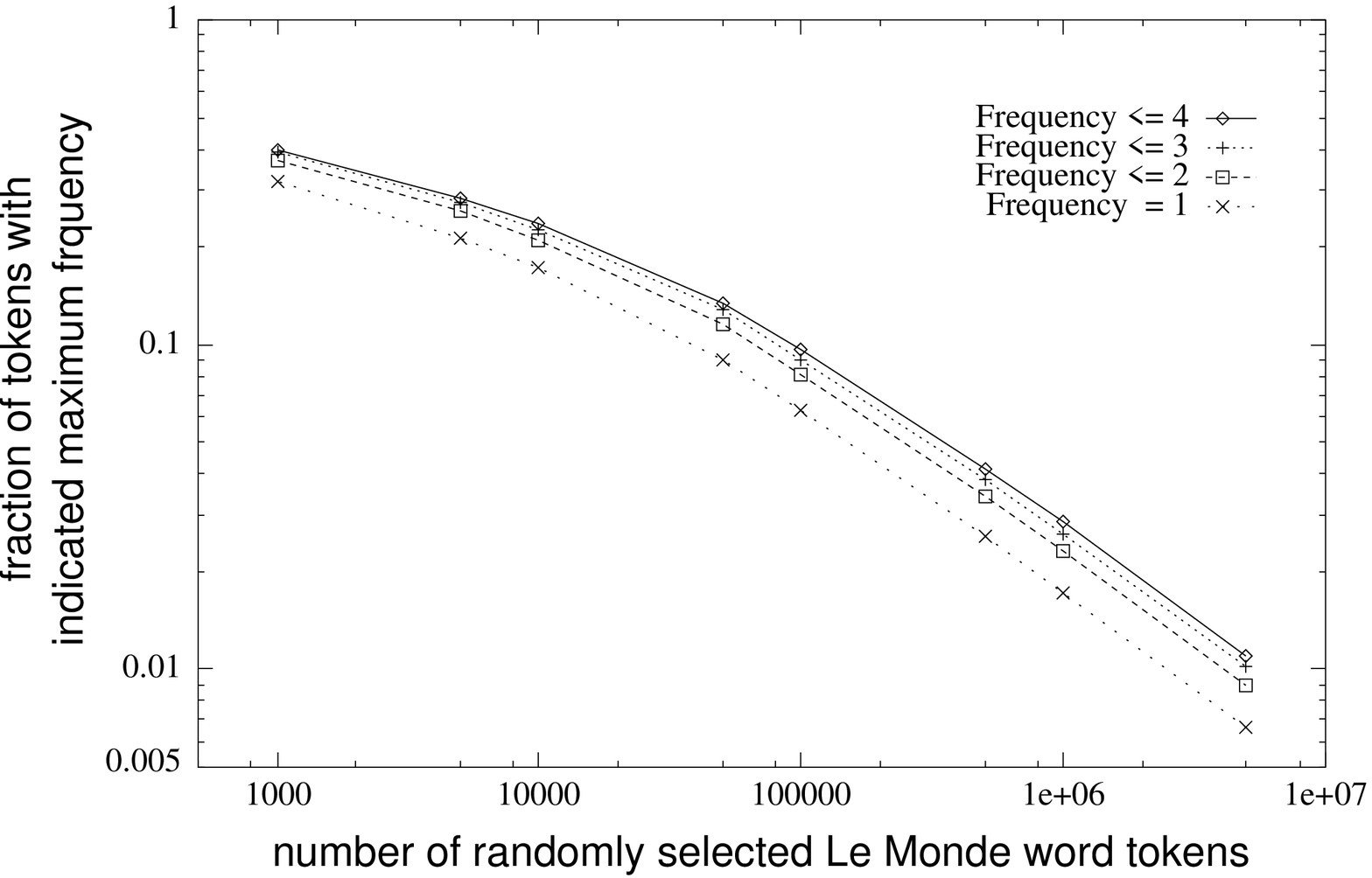,width=5.25in}}
\caption[{\em The log-log relationship for singletons and higher
frequencies}]{{\em The log-log relationship for higher frequencies.
The bottom curve in this graph is the same as the top curve in
Figure~\ref{singmass}}\label{frqmass}}
\end{figure}

\begin{figure}[htb]
\centerline{\psfig{figure=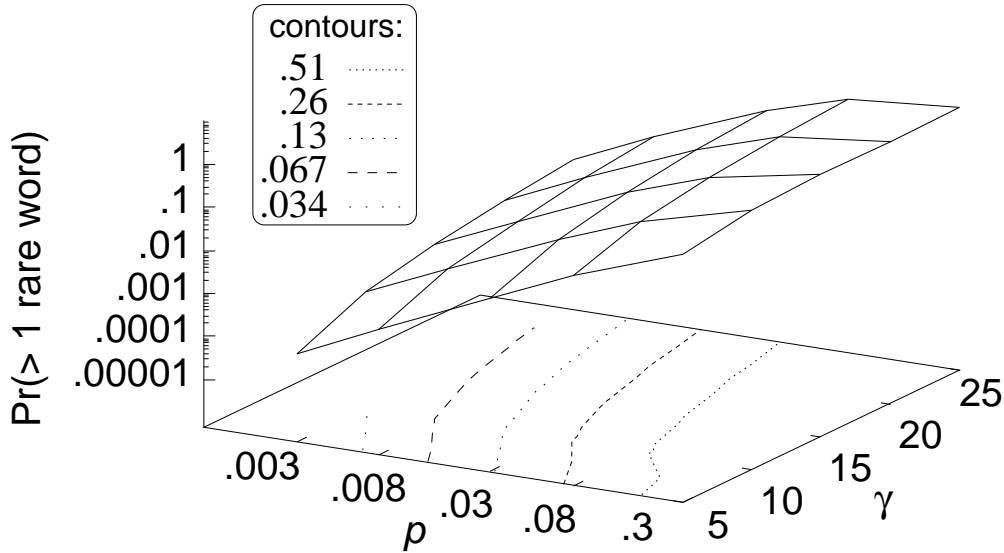,width=5.25in}}
\caption{{\em Probability of more than one rare word
co-occurring.}\label{rare}}
\end{figure}

Corpus size determines the probability that a randomly chosen word
will have a particular frequency.  The likelihood of a correct link
for a rare word token $w$ also depends on one other variable.  If $w$
co-occurs with only one rare word (in the opposite half of the
bitext), then the competitive linking algorithm is likely to eliminate
all of $w$'s indirect associations before it attempts to link $w$.
Problems arise only when more than one candidate remains for linking
to $w$.  What is the probability that $w$ co-occurs with more than one
rare word?  The analysis is easiest under the distance-based model of
co-occurrence, where the threshold $\delta$ on the distance from the
bitext map is specified in words rather than in characters
\citep{coocmod}.  Suppose that $w$ co-occurs with $\gamma$ words in
the opposite half of the bitext, where $\gamma$ is either the vertical
or horizontal component of $\delta$\footnote{{\em I.e.}\ $\gamma$ is
the same as \citet{wordalign}'s window width.}.  Let $p$ be the
probability that a word co-occurring with $w$ is rare.  Then the
probability of exactly $k$ rare words co-occuring with $w$ can be
approximated by a binomial distribution with parameters $\gamma$ and
$p$.  It follows that the probability of more than one rare word
co-occurring with $w$ is
\begin{equation}
\label{2sing}
\Pr(> \mbox{1 rare word co-occuring}) = 1 - B( 0 | \gamma, p) - B( 1 |
\gamma, p) .
\end{equation}
Figure~\ref{rare} plots Equation~\ref{2sing} over different values of
$\gamma$ and $p$.  The range of $p$ corresponds roughly to the range
of the y-axis in Figures~\ref{singmass} and~\ref{frqmass}.  The figure
illustrates how the power of the one-to-one assumption varies with
corpus size.

\newpage

\section{Evaluation}
\label{TM-eval}

\subsection{Evaluation By Token}
\label{bytoken}

This section compares translation model estimation methods A, B and C
to each other and to \citet{ibm}'s Model~1.  Until now, translation
models have been evaluated either subjectively \citep[{\em e.g.}\
][]{arpa} or using relative metrics, such as perplexity with respect
to other models \citep{ibm}.  More objective and more accurate tests
can be carried out using a ``gold standard.'' I hired bilingual
annotators to link roughly sixteen thousand corresponding words
between on-line versions of the Bible in French and English.  This
bitext was selected to facilitate widespread use and standardization
\citep[see][for details]{blinker}.  The entire Bible bitext comprised
29614 verse pairs, of which 250 verse pairs were hand-linked using a
specially developed annotation tool.  The annotation style guide
\citep{styleguide} was based on the intuitions of the annotators, so
it was not biased towards any particular translation model.  The
annotation was carried out 5 times by different annotators.

A straightforward metric for evaluating a translation model with
respect to a gold standard can be derived from the recall and
precision measures widely used in the information retrieval
literature.  When
comparing a set of ``test'' elements $X$ to a set of ``correct''
elements~$Y$,
\begin{equation}
\label{precision}
precision(X | Y) = \frac{|X \cap Y|}{|X|} ,
\end{equation}
\begin{equation}
\label{recall}
recall(X | Y) = \frac{|X \cap Y|}{|Y|} .
\end{equation}
$X$ and $Y$ can be fuzzy sets, such as probability distributions, in
which case $|X|$ is defined as the sum of the weights of the elements
in $X$ and $|X \cap Y|$ is the sum of the weights of the elements
shared by $X$ and $Y$.

Equations~\ref{precision} and~\ref{recall} differ only in the set
whose size is used as the denominator.  If neither $X$ nor $Y$ is
privileged, or if precision and recall are equally important, we can
compute a symmetric measure of agreement $D$ as the harmonic mean of
precision and recall:
\begin{eqnarray}
\label{setDice}
D(X,Y) = & \frac{1}{\frac{1}{Precision(X | Y)} + \frac{1}{Recall(X |
Y)}} = & \frac{2 \cdot |X \cap Y|}{|X| + |Y|} .
\end{eqnarray}
$D$ is the set-theoretic equivalent of the Dice coefficient
\citep{dice} and conveniently ranges from zero to one.  

To reiterate, Model~1 is based on co-occurrence information only;
Method~A is based on the one-to-one assumption; Method~B adds the ``one
sense per collocation'' hypothesis to Method~A; Method~C conditions the
auxiliary parameters of Method~B on various word classes.  Whereas
Methods~A and B and Model~1 were fully specified in
Section~\ref{cooconly} and Section~\ref{paramest}, the latter section
described a variety of features on which Method~C might classify
words.  For the purposes of the experiments reported in this article,
Method~C employed the simple classification in
Table~\ref{methodCclasses} for both languages in the
\begin{table*}[htb]
\centering
\begin{tabular}{|c|l|}
\hline
Class Code & Description \\ \hline
EOS & End-Of-Sentence punctuation \\
EOP &  End-Of-Phrase punctuation, such as commas and colons \\
SCM & Subordinate Clause Markers, such as '' and ( \\
SYM & Symbols, such as \~{} and $\ast$ \\
NU & the {\sc null} word, in a class by itself \\
C & Content words: nouns, adjectives, adverbs, non-auxiliary verbs \\
F & all other words, {\em i.e.}\ function words \\
\hline
\end{tabular}
\caption[{\em Word classes used by Method~C}]{{\em Word classes used
by Method~C for the experiments reported in this
article.}\label{methodCclasses}}
\end{table*}
bitext.  All classification was performed by table lookup; no
context-aware part-of-speech tagger was used.  In particular, words
that were ambiguous between open classes and closed classes were
always deemed to be in the closed class.  The only language-specific
knowledge involved in this classification method is the list of
function words in class F.  Certainly, more sophisticated word
classification methods could produce better models, but even the
simple classification in Table~\ref{methodCclasses} should suffice to
demonstrate the method's potential.

Each of the four methods was used to estimate a word-to-word
translation model from the 29614 verse pairs in the Bible bitext.  All
methods were deemed to have converged when less than .0001 of the
translational probability distribution changed from one iteration to
the next.  The links assigned by each of methods A, B and C in the
last iteration were normalized into joint probability distributions
using Equation~\ref{mletrans}.  I shall refer to these joint
distributions as Model~A, Model~B and Model~C, respectively.  Each of
the joint probability distributions was further normalized into two
conditional probability distributions, one in each direction.  Since
Model~1 is inherently directional, its conditional probability
distributions were estimated separately in each direction, instead of
being derived from a joint distribution.

The four models' predictions were compared to the gold standard
annotations.  Although the models were evaluated on part of the same
bitext on which they were trained, the evaluations were with respect
to the translational equivalence relation hidden in this bitext, not
with respect to any of the bitext's visible features.  Such testing on
training data is acceptable for unsupervised learning algorithms.

\begin{figure}[htb]
\centerline{\psfig{figure=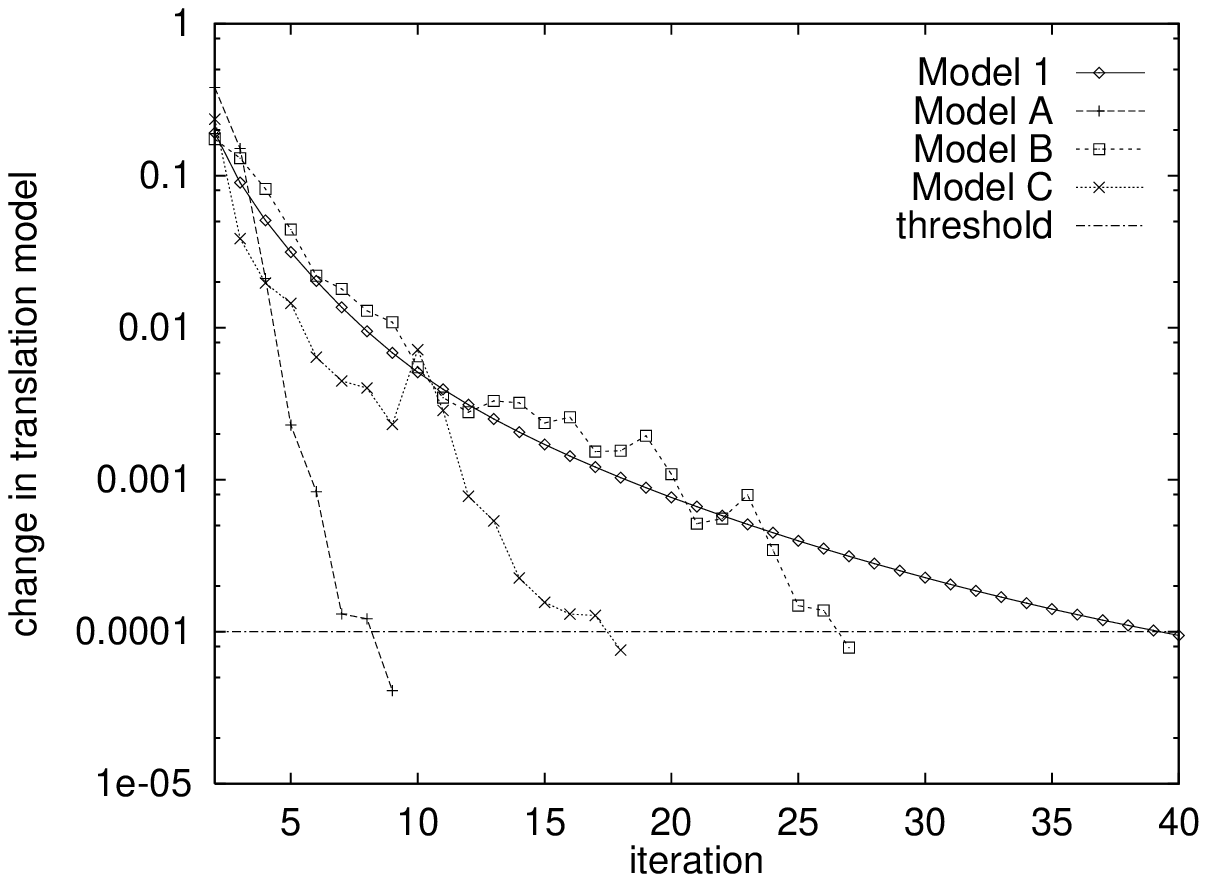,width=5.25in}}
\caption[{\em Convergence rates for Model~1 and Methods~A, B,
and~C}]{{\em Convergence rates for Model~1 and Methods~A, B, and~C.
Changes from each iteration to the next were measured in terms of the
set-theoretic Dice coefficient.}\label{convergence}}
\end{figure}

Before comparing the accuracies of the different models, it is
interesting to consider their convergence rates.
Figure~\ref{convergence} shows that, although the EM algorithm
guarantees monotonic convergence for Model~1, it requires more
iterations to converge on these training data than models A, B and~C.
To be fair, we must remember that Method~B and Method~C take time to
estimate their auxiliary parameters on each iteration.  So,
Figure~\ref{convergence} does not say which method is fastest in real
time.  Such a comparison is very dependent on the details of each
method's implementation.  In the current (very inefficient)
implementations, Model~A converged in about 6 hours, Model~B in about
20 hours, Model~C in about 24 hours and Model~1 converged in about 27
hours.

\begin{figure}[H]
(a) \linebreak
\centerline{\psfig{figure=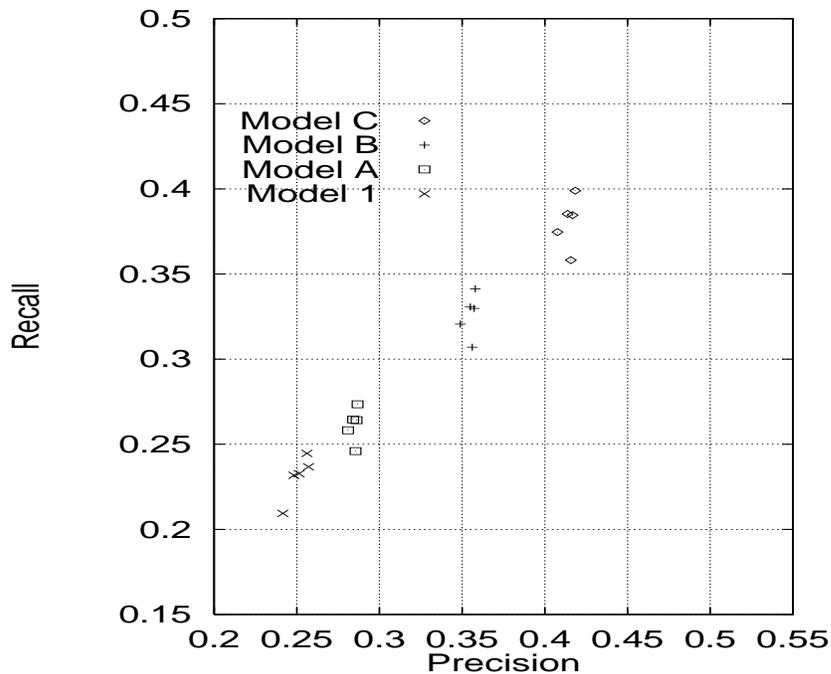,height=3.5in,width=4.5in}} \linebreak
(b) \linebreak
\centerline{\psfig{figure=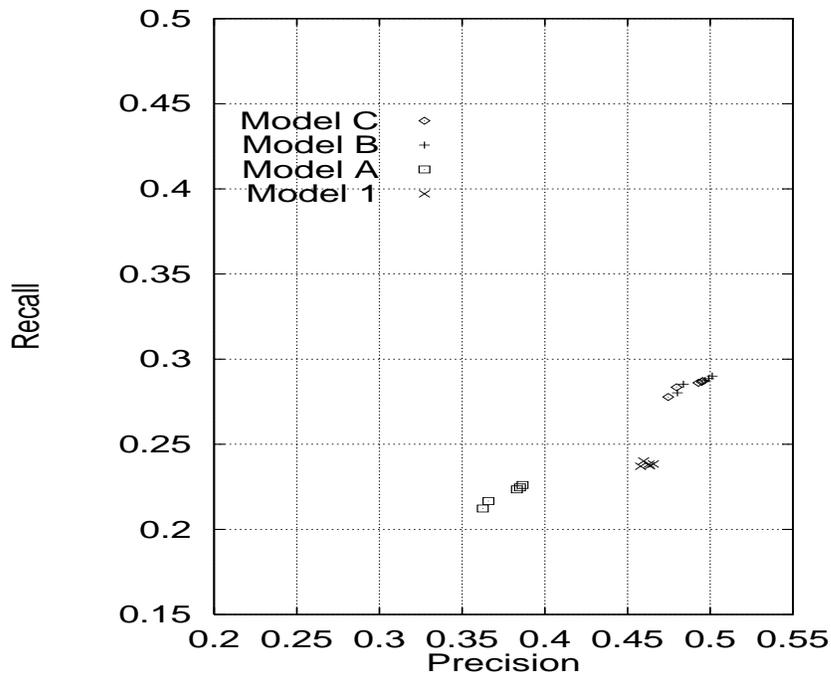,height=3.5in,width=4.5in}}
\caption[{\em Comparison of model performance on ``single-best'' translation
task}]{{\em Comparison of model performance on ``single-best''
translation task. (a) All links; (b) open-class links
only.}\label{top}}
\end{figure}

The first evaluation was on ``single-best'' translation of the
kind that somebody might use to get the gist of a foreign-language
document.  The input to the experiment was one side of the gold standard
bitext.  The output was the model's single best guess about the
translation of each word in the input, together with the input word.
In other words, each model produced link tokens consisting of input
words and their translations.  I computed the models' precision and
recall by comparing the link tokens produced by each model to the link
tokens in the gold standard.  The accuracy of each model was averaged
over the two directions of translation: English to French and French
to English.  Figure~\ref{top}(a) shows that each of the innovations
introduced in Section~\ref{paramest} improves both precision and
recall on this task on these data.  The gold standard bitext was
actually annotated five times by seven different annotators.  This
replication helped to establish statistical significance among the
differences in model accuracy.  The performance differences reported
in this section are statistically significant at the $\alpha = .05$
level, according to the Wilcoxon signed ranks test.

Some applications don't care about function words.  To get a sense of
the relative effectiveness of the different translation model
estimation methods when function words are taken out of the equation,
I removed all closed-class words (including non-alphabetic
symbols) from the models and renormalized the conditional
probabilities.  Then, I removed from the gold standard all link
tokens where one or both of the linked words were closed-class words.
Finally, I recomputed precision and recall.  The results are shown in
Figure~\ref{top}(b).  When closed-class words were ignored, Model~1
performed better than Method~A, because open-class words are more
likely to violate the one-to-one assumption.  However, the explicit
error model in Methods~B and~C boosted their recall and precision
significantly higher than Model~1 and Method~A.  As expected, there was
no significant difference in accuracy between Method~B and Method~C on
this task, because it left only two classes for Method~C to
distinguish: content words and {\sc null}s.

\begin{figure}[H]
(a) \linebreak
\centerline{\psfig{figure=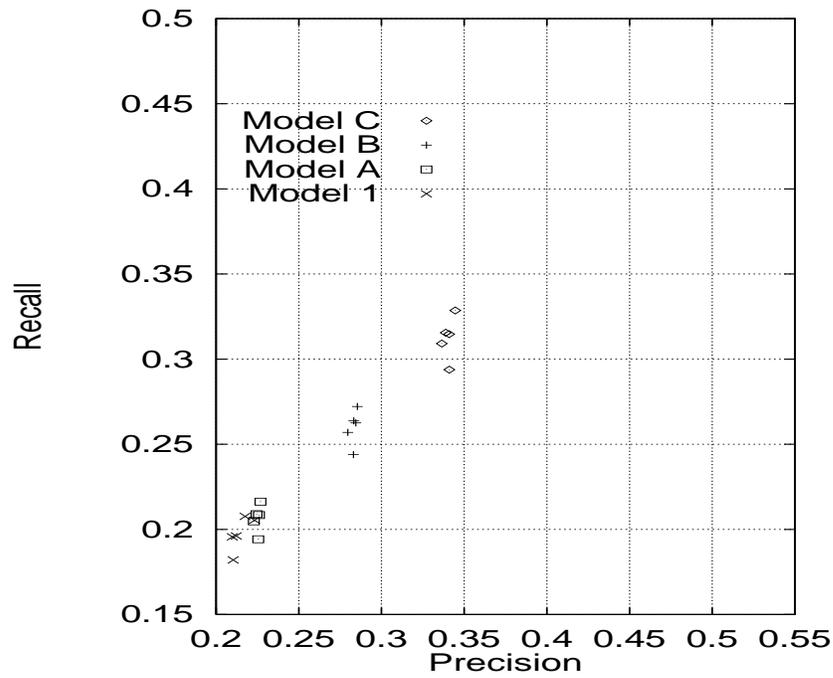,height=3.5in,width=4.5in}} \linebreak
(b) \linebreak
\centerline{\psfig{figure=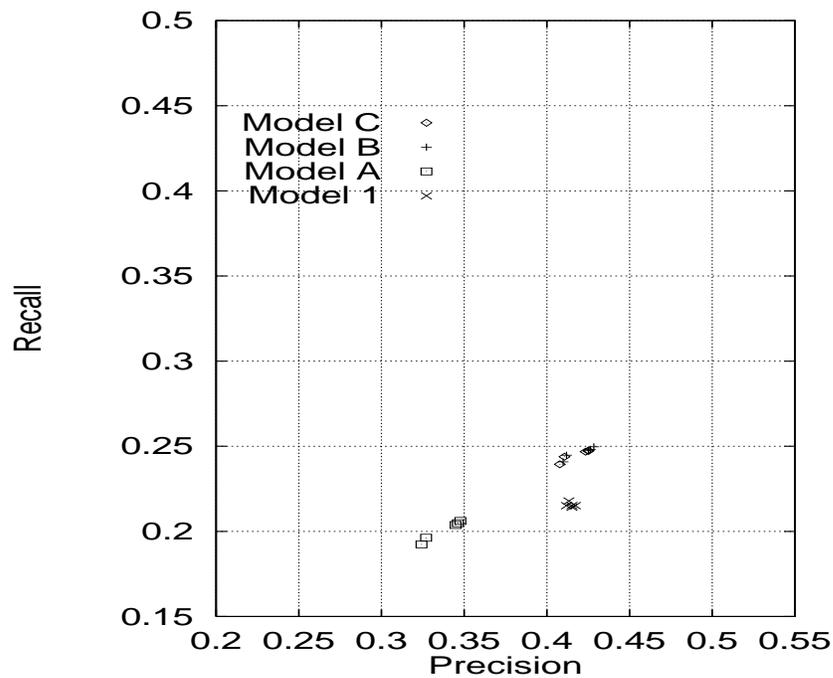,height=3.5in,width=4.5in}}
\caption[{\em Comparison of model performance on ``whole
distribution'' task}]{{\em Comparison of model performance on ``whole
distribution'' task. (a) All links; (b) open-class links
only.}\label{dist}}
\end{figure}

For some applications, it is insufficient to guess only the most
likely translation of each word in the input.  The model is expected
to output the entire distribution of possible translations for each
input word.  This distribution is then convolved with other
distributions that are relevant to the application.  For example, in
cross-language information retrieval, the translational distribution
is convolved with the distribution of term frequencies.  In
statistical machine translation, the translational distribution can be
convolved with a target language model \citep{candide}.  To see how
the different models might perform on this ``whole distribution''
task, I performed a second set of experiments.  This time, the models
generated a whole set of links from each input word, weighted
according to the probability assigned to each of the input word's
translations.  I computed the precision and recall of the fuzzy sets
of links generated by the models to the five gold standard annotations
as before.  I repeated the experiment once with closed-class words and
once without, and again averaged the results over the two directions
of translation.  The results are in Figure~\ref{dist}, which is
plotted on the same scale as Figure~\ref{top} to facilitate
comparison.  The only change in the relative accuracy of the models was
that Methods~B and~C no longer had significantly higher precision than
Model~1 when closed-class words were ignored.
However, all the scores were lower than their counterparts on the
``single-best'' translation task, because it is more difficult for any
statistical method to correctly model the less common translations.
The ``best'' translations are usually the most common.

\begin{figure}[htb]
\centerline{\psfig{figure=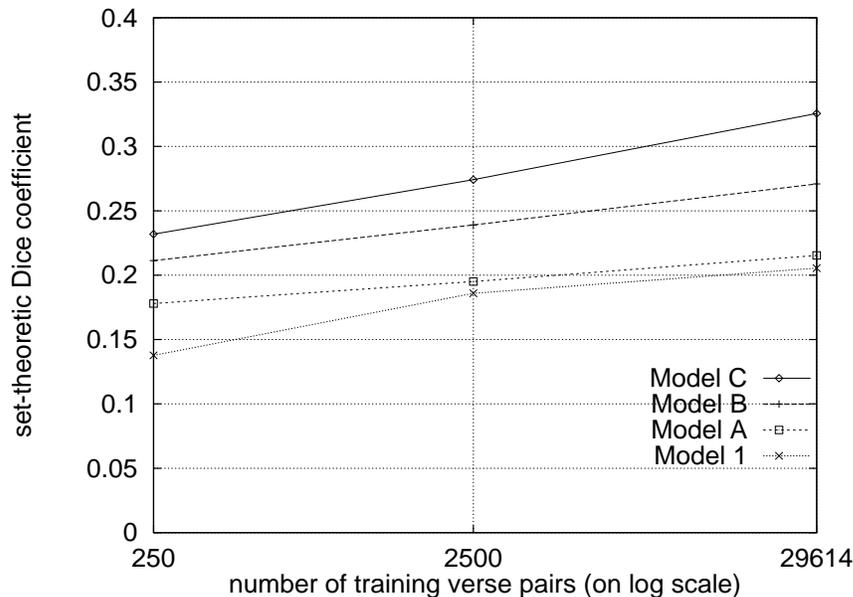,width=4.5in}}
\caption[{\em Effects of training set size on model
accuracy}]{{\em Effects of training set size on model accuracy
on the ``whole distribution'' task.}\label{bysize}}
\end{figure}

\newpage
To study how the benefits of the various biases vary with training
corpus size, I evaluated Models~A, B, C and~1 on the ``whole
distribution'' translation task, after training them on three
different-size subsets of the Bible bitext.  The first subset
consisted of only the 250 verse pairs in the gold standard.  The
second subset included these 250 plus another random sample of 2250
for a total of 2500, an order of magnitude larger than the first
subset.  The third subset contained all 29614 verse pairs in the Bible
bitext, roughly an order of magnitude larger than the second subset.
All models were compared to the five gold standard annotations.  The
correlation between recall and precision was very high on this task
($\rho = .99$), as illustrated in Figure~\ref{dist}(a).  So, the
results can be well represented by the set-theoretic Dice coefficient
in Equation~\ref{setDice}, as applied to probabilistic (fuzzy) sets.
The mean Dice scores over the five gold standard annotations are
graphed in Figure~\ref{bysize}.
The figure suggests that, at least for French/English translation
models, each of the biases presented in this article improves the
efficiency of modeling the available training data.  The one-to-one
assumption is useful, even though it is tractable only under a greedy
estimation method.  In relative terms, the advantage of the one-to-one
assumption is much more pronounced on smaller training sets.  For
example, Model~A is 29\% more accurate than Model~1 when trained on
only 250 verse pairs.  The explicit error model buys a considerable
gain in accuracy across all sizes of training data, as do the link
classes of Model~C.  In concert, on the gold standard test set, the
three biases outperformed Model~1 by up to 55\%.  This difference is
even more significant given the absolute performance ceiling of 82\%
established by the inter-annotator agreement rates on the gold
standard \citep{blinker}.

\subsection{Evaluation By Type}
\label{bytype}

An important application of statistical translation models is to help
lexicographers compile bilingual dictionaries.  Dictionaries are
written to answer the question, ``What are the possible translations
of X?''  This is a question about link types, rather than about link
tokens.

Evaluation by link type is a thorny issue.  Human judges often
disagree about the degree to which context should play a role in
judgements of translational equivalence.  For example, the
Harper-Collins French Dictionary \citep{newdict} gives the following
French translations for English {\em appoint: nommer, engager, fixer,
d\'{e}signer}.  Likewise, most lay judges would not consider {\em
instituer} a correct French translation of {\em appoint}.  In actual
translations, however, when the object of the verb is {\em commission,
task force, panel}, {\em etc.}, English {\em appoint} is usually
translated into French as {\em instituer}.  To account for this kind
of context-dependent translational equivalence, link types must be
evaluated with respect to the bitext whence they were induced.  

I performed a post-hoc evaluation of the link types produced by an
earlier version of Method~B.  The bitext used for this evaluation was
the same aligned Hansards bitext used by \citet{wordcorr}, except that
I used only 300,000 aligned segment pairs to save time.  The bitext
was automatically pre-tokenized to delimit punctuation, English
possessive pronouns and French elisions.  Morphological variants in
both halves of the bitext were stemmed to a canonical form.
 
\begin{figure}[htb]
\centerline{\psfig{figure=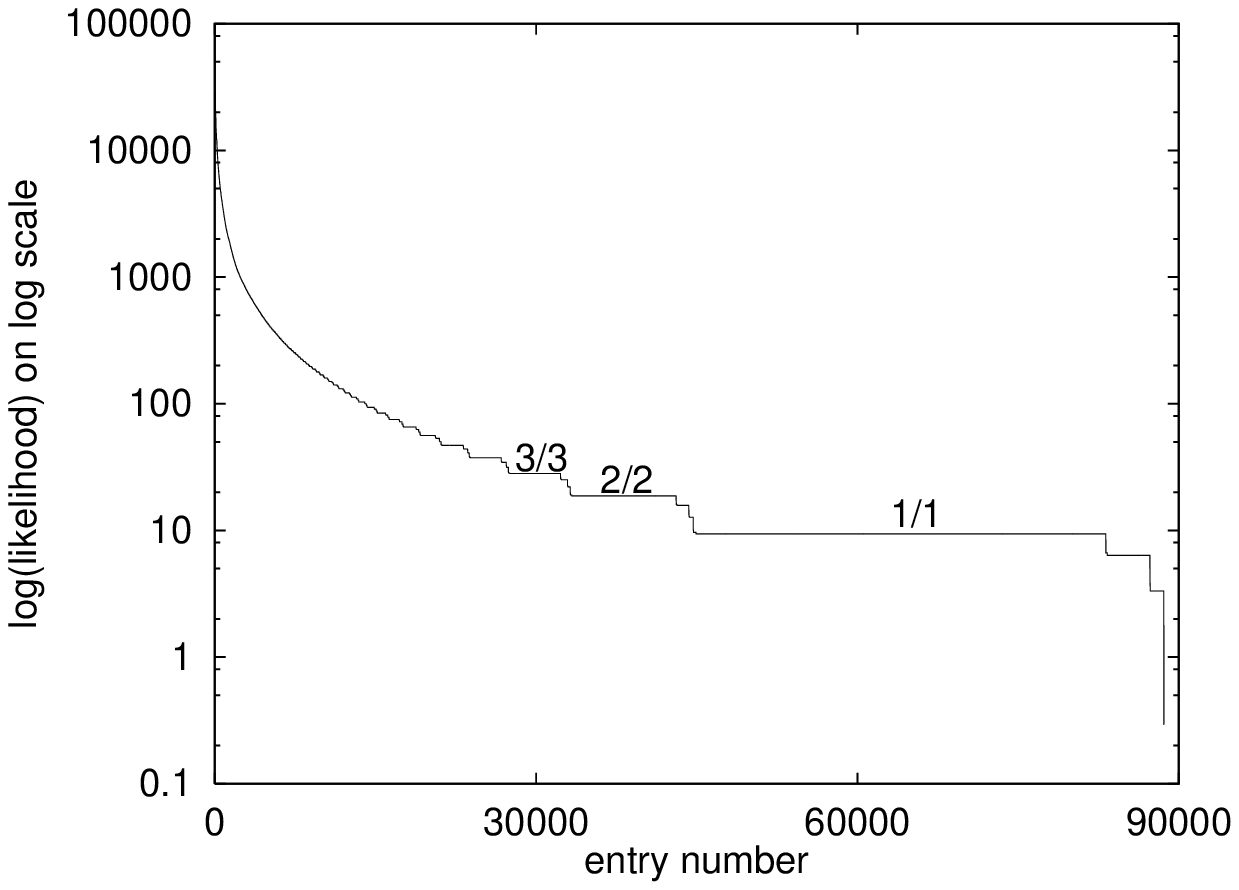,width=5in}}
\caption[{\em Distribution of link type log-likelihood scores}]{{\em
Distribution of link type log-likelihood scores.  The long plateaus
correspond to the most common combinations of $\frac{links({\bf
u,v})}{cooc({\bf u,v})}: 1/1, 2/2 \mbox{ and } 3/3$.}\label{curve}}
\end{figure}

The link types assigned by the converged model were sorted by the
log-likelihood scores in Equation~\ref{ll}.  Figure~\ref{curve} shows
the distribution of these scores on a log scale.  The log scale helps
to illustrate the plateaus in the curve.  The longest plateau
represents the set of word pairs that were linked once out of one
co-occurrence (1/1) in the bitext.  All these word pairs were equally
likely to be correct.  The second-longest plateau resulted from word
pairs that were linked twice out of two co-occurrences (2/2) and the
third longest plateau is from word pairs that were linked three times
out of three co-occurrences (3/3).  As usual, the entries with higher
likelihood scores were more likely to be correct.  By discarding
entries with lower likelihood scores, recall could be traded off for
precision.  This trade-off was measured at three points, representing
cutoffs at the end of each of the three longest plateaus.

The traditional method of measuring recall requires knowledge of the
correct link types, which is impossible to determine without a gold
standard.  An approximate recall measure can be based on the number of
different words in the corpus.  For lexicons extracted from corpora,
perfect recall implies at least one entry containing each word in the
corpus.  One-sided variants, which consider only source words, have
also been used \citep{wordcorr}.
\begin{table*}[htb]
\centering
\caption[{\em Lexicon recall at three different minimum likelihood
thresholds}]{{\em Lexicon recall at three different minimum
likelihood thresholds.  The bitext contained 41,028 different English
words and 36,314 different French words, for a total of 77,342.}}
\label{recall-bytype}
\begin{tabular}{|c|c|c|cc|cc|}
\hline
 & minimum & total & English & & French & \\
cut-off & likelihood& lexicon & words & & words & \\
plateau & score & entries & represented & \% & represented & \% \\
\hline
3/3 & 28 & 32274 & 14299 & 35 & 13409 & 37 \\
2/2 & 18 & 43075 & 18533 & 45 & 17133 & 47 \\
1/1 & 9 & 88633 & 36371 & 89 & 33017 & 91 \\
\hline
\end{tabular}
\end{table*}
Table~\ref{recall-bytype} reports both the marginal (one-sided) and
the combined recall at each of the three cut-off points.  It also
reports the absolute number of (non-{\sc null}) entries in each of the
three lexicons.  Of course, the size of automatically induced lexicons
depends on the size of the training bitext.  Table~\ref{recall-bytype}
shows that, given a sufficiently large bitext, the method can
automatically construct translation lexicons with as many entries as
published bilingual dictionaries.

The next task was to measure precision.  It would have taken too long
to evaluate every lexicon entry manually.  Instead, I took 5 random
samples (with replacement) of 100 entries each from each of the three
lexicons.  Each of the samples was first compared to a translation
lexicon extracted from a machine readable bilingual dictionary
\citep{collins}.  All the entries in the sample that appeared in the
dictionary were assumed to be correct.  I checked the remaining
entries in all the samples by hand.  To account for context-dependent
translational equivalence, I evaluated the precision of the
translation lexicons in the context of the bitext whence they were
extracted, using a simple bilingual concordancer.  A lexicon entry
{\bf (u,v)} was considered correct if $u$ and $v$ ever appeared as
direct translations of each other in an aligned segment pair.

Direct translations come in different flavors.  Most entries that
I checked by hand were of the plain vanilla variety that you might
find in a bilingual dictionary (entry type V).  However, a significant
number of words translated into a different part of speech (entry
type P).  For instance, in the entry (protection, prot\'{e}g\'{e}),
the English word is a noun but the French word is an adjective.  This
entry appeared because ``to have protection'' is often translated as
``\^{e}tre prot\'{e}g\'{e}'' in the bitext.  The entry will never
occur in a bilingual dictionary, but users of translation lexicons, be
they human or machine, will want to know that translations often
happen this way.  Incomplete entries, described above, were counted in
a third category (entry type I).

\begin{figure}[H]
\centerline{\psfig{figure=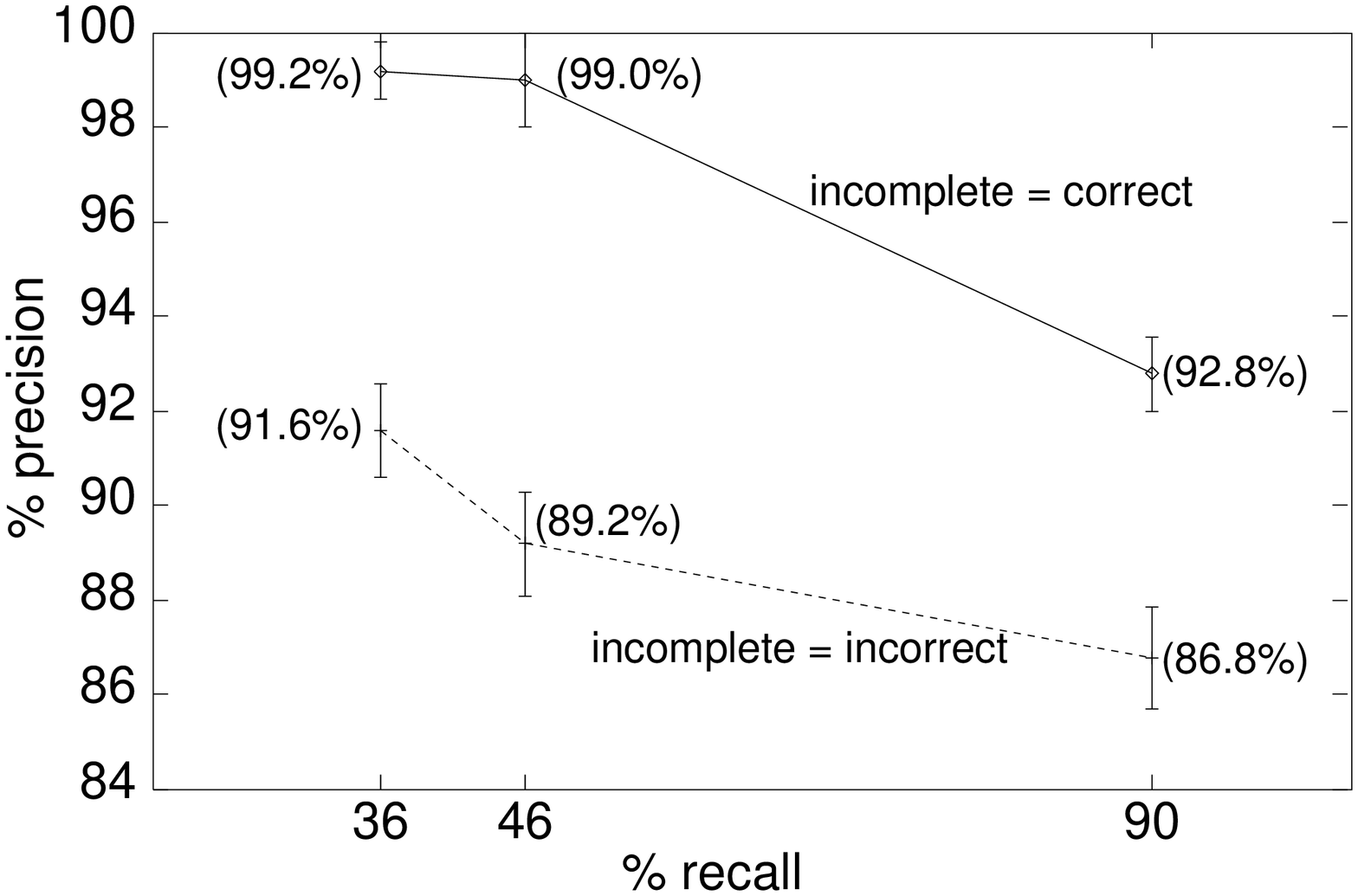,width=5.25in}}
\caption[{\em Translation lexicon precision at varying levels of
recall}]{{\em Translation lexicon precision with 95\% confidence
intervals at varying levels of recall.}}
\label{prec-bytype}
\end{figure}

\begin{table*}[htb]
\centering
\caption[{\em Distribution of different types of correct lexicon
entries.}]{{\em Distribution of different types of correct lexicon
entries at varying levels of recall \mbox{(mean $\pm$ standard deviation)}.}}
\label{cats}
\begin{tabular}{|c|c||c|c|c||c|}
\hline
cutoff & recall & \% type V & \% type P & \% type I & total \% precision \\
\hline
3/3 & 36\% & 89 $\pm$ 2.2 & 3.4 $\pm$ 0.5 & 7.6 $\pm$ 3.2 & 99.2 $\pm$ 0.8 \\
2/2 & 46\% & 81 $\pm$ 3.0 & 8.0 $\pm$ 2.1 & 9.8 $\pm$ 1.8 & 99.0 $\pm$ 1.4 \\
1/1 & 90\% & 82 $\pm$ 2.5 & 4.4 $\pm$ 0.5 & 6.0 $\pm$ 1.9 & 92.8 $\pm$ 1.1 \\
\hline
\end{tabular}
\end{table*}
Table~\ref{cats} reports the distribution of correct lexicon entries
among the types V, P and~I.  Figure~\ref{prec-bytype} graphs the
precision of the method against recall, with 95\% confidence
intervals.  The upper curve represents precision when incomplete
links are considered correct, and the lower when they are considered
incorrect.  On the former metric, the method can generate translation
lexicons with precision and recall both exceeding 90\%, as well as
dictionary-sized translation lexicons that are over 99\% correct.

\section{Conclusion}

There are many ways to model translational equivalence and many ways
to estimate translation models.  ``The mathematics of statistical
machine translation'' proposed by \citet{ibm} are just one kind of
mathematics for one kind of statistical translation.  In this article,
I have proposed and evaluated new kinds of translation model biases,
alternative parameter estimation strategies, and general techniques
for exploiting pre-existing knowledge that may be available about
particular languages and language pairs.  On a variety of evaluation
metrics, each infusion of knowledge about the problem domain resulted
in better translation models.

Each innovation presented here opens the way for more research.  Model
biases can be mixed and matched with each other, with previously
published biases like the word order correlation bias, and with other
biases yet to be invented.  The competitive linking algorithm can be
generalized in various ways.  New kinds of pre-existing knowledge can
be exploited to effect significant accuracy improvements for
particular language pairs or even just for particular bitexts.  It is
difficult to say where the greatest advances will come from.  Yet, one
thing is clear from our current vantage point: Research on empirical
methods for modeling translational equivalence has not run out of
steam, as some have claimed, but has only just begun.

\section*{Acknowledgements}
Many of the ideas in this paper came from enlightening correspondence
with Ken Church, Mike Collins, Ido Dagan, Jason Eisner, Steven Finch,
George Foster, Djoerd Hiemstra, Adwait Ratnaparkhi and Lyle Ungar.
This research was supported by DARPA grant N6600194C-6043.


\end{document}